\documentclass[runningheads]{llncs}

\usepackage{eccv}

\usepackage{eccvabbrv}

\usepackage{graphicx}
\usepackage{booktabs}
\usepackage{multirow}
\usepackage{bm}
\usepackage[accsupp]{axessibility}  
\newcommand{\R}{\mathbb{R}}

\newcommand{\x}{\mathbf{x}}

\newcommand{\zv}{\mathbf{z}}

\newcommand{\tv}{\mathbf{t}}
\newcommand{\Rv}{\mathbf{R}}

\newcommand{\mesh}{\mathcal{M}}

\usepackage{hyperref}

\usepackage{orcidlink}

\begin{document}

\title{Physically Grounded 3D Generative
Reconstruction under Hand Occlusion using
Proprioception and Multi-Contact Touch}

\titlerunning{Physically Grounded 3D Generative
Reconstruction}

\author{Gabriele M. Caddeo\inst{1}\orcidlink{0000-0003-1566-1314} \and
Pasquale Marra\inst{1}\orcidlink{0009-0002-7110-1952} \and
Lorenzo Natale\inst{1}\orcidlink{0000-0002-8777-5233}}

\authorrunning{G. M. ~Caddeo et al.}

\institute{
Istituto Italiano di Tecnologia\\
\email{\{gabriele.caddeo,pasquale.marra,lorenzo.natale\}@iit.it}\\
\url{https://hsp.iit.it/}
}
\maketitle
\vspace{-0.2in}
\begin{center}
    \centering
    \includegraphics[width=\linewidth]{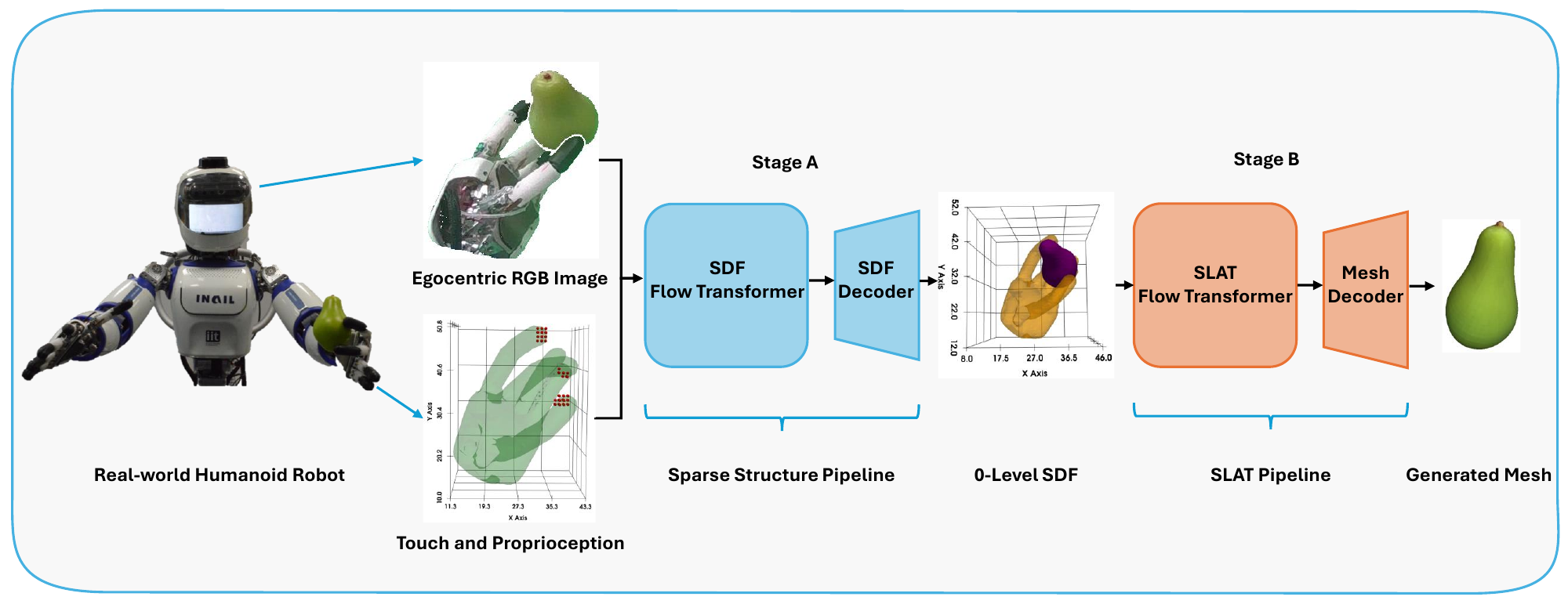}
    \captionof{figure}{\textbf{Inference pipeline}. We present a method for physically plausible 3D shape reconstruction by fusing vision and contact. Specifically, we use active contact information from tactile sensors distributed on the hand, and negative information, i.e., non-interpenetration, from the hand geometry. In addition, the method requires an egocentric RGB image of the object, together with the hand and object masks. A Flow Transformer conditioned on multiple senses then generates a physically consistent SDF on a 3D grid in the same geometric domain as the hand. Finally, a Structured Latents (\textbf{SLat}) pipeline can generate a refined textured mesh, directly at metric scale.}
    \label{fig:first_image}
    \vskip -4,5em
\end{center}
\vspace{-0.15in}
\begin{abstract}
  We propose a multimodal, physically grounded approach for metric-scale amodal object reconstruction and pose estimation under severe hand occlusion. Unlike prior occlusion-aware 3D generation methods that rely only on vision, we leverage physical interaction signals: proprioception provides the posed hand geometry, and multi-contact touch constrains where the object surface must lie, reducing ambiguity in occluded regions. We represent object structure as a pose-aware, camera-aligned signed distance field (SDF) and learn a compact latent space with a Structure-VAE. In this latent space, we train a conditional flow-matching diffusion model, pretraining on vision-only images and finetuning on occluded manipulation scenes while conditioning on visible RGB evidence, occluder/visibility masks, the hand latent representation, and tactile information. Crucially, we incorporate physics-based objectives and differentiable decoder-guidance during finetuning and inference to reduce hand–object interpenetration and to align the reconstructed surface with contact observations. Because our method produces a metric, physically consistent structure estimate, it integrates naturally into existing two-stage reconstruction pipelines, where a downstream module refines geometry and predicts appearance. Simulation experiments show that adding proprioception and touch substantially improves completion under occlusion and yields physically plausible reconstructions at correct real-world scale compared to vision-only baselines; we further validate transfer by deploying the model on a real humanoid robot with an end-effector different from those used during training. See \url{https://github.com/hsp-iit/physical-generative-reconstruction}
\end{abstract}

\vspace{-0.29in}
\section{Introduction}
\label{sec:intro}
Accurate object geometry is important to guide robotic manipulation. This includes  knowledge of the object surface to compute stable contacts, free space for collision avoidance, and  object shape to plan subsequent actions. Yet the moments when geometry matters most are also when vision is least reliable. During grasping and in-hand manipulation, the hand severely occludes the object, and single-view RGB reconstruction becomes fundamentally underconstrained. Vision-only reconstructions may look plausible but remain unusable for control, e.g., intersecting the hand, missing true contact regions, or drifting in metric scale. Manipulation naturally produces \emph{physical evidence}. Contacts and known hand kinematics provide useful cues to resolve the ambiguities caused by occlusion. Tactile sensing has long been used to estimate object properties~\cite{10851808,8460494}, pose~\cite{10160359}, and to improve robustness of grasping~\cite{11127723, jiang2022shallitouchvisionguided} and in-hand manipulation~\cite{9811953, oller2022manipulationmembraneshighresolutionhighly}, and recent advances in tactile hardware and learning-based models have made touch increasingly practical as a perception signal. Yet these cues are rarely integrated as \emph{geometric constraints} for dense 3D reconstruction under occlusion. Critically, contact provides direct constraints on geometry exactly where vision is missing: observed touches indicate where the surface must be, while physical feasibility (\eg, non-interpenetration with the end-effector) constrains where the object cannot be.

Despite their relevance, such interaction cues are still not commonly integrated as explicit geometric constraints in dense 3D reconstruction under severe occlusion.
In parallel, recent diffusion and flow-based 3D generative models have improved single-view 3D inference by learning strong shape priors, but they are typically optimized for visual plausibility and are rarely grounded in physical constraints~\cite{li2025dso}. Under heavy occlusion, this can produce samples that violate contact feasibility or drift in scale, limiting their impact for downstream robotic tasks (\eg, insertion~\cite{tang2024automatespecialistgeneralistassembly} or pose estimation~\cite{wen2024foundationposeunified6dpose}). 

We address this gap by treating in-hand 3D reconstruction as a \textbf{physically grounded} generative inference problem. Building upon~\cite{xiang2024structured}, we propose a multimodal approach for metric-scale amodal object reconstruction under severe hand occlusion from a single egocentric RGB observation, proprioceptive hand pose, and multi-contact tactile feedback. Our key contribution operates at the structure generation stage: we represent object geometry as a pose-aware, camera-aligned signed distance field (SDF), providing a continuous and differentiable representation that enables gradient-based physical constraints. We learn a compact latent space for SDF grids with a Structure-VAE and train a conditional flow-matching model to infer this latent. The model is pretrained on vision-only object images to learn a strong shape prior, then finetuned on occluded manipulation scenes while conditioning on visible RGB evidence and masks, a latent encoding of the posed hand, and a learned volumetric touch representation. To enforce physical validity, we incorporate two interaction-derived objectives: a contact-consistency loss that pulls the reconstructed surface toward observed contacts, and a non-interpenetration loss that pushes the object out of the hand interior. We further use these same terms as decoder-based physics guidance during sampling, steering generation toward shapes that satisfy contact and feasibility constraints. Since we reconstruct the object in the same 3D domain as the end-effector, the pose is implicitly estimated.
Moreover, our method integrates naturally into state-of-the-art shape–appearance pipelines: the physically consistent structure can be passed to a downstream refinement/texturing module (made pose- and occlusion-aware) to obtain textured reconstructions consistent with the camera view and the occluding hand. Alternatively, our structure estimate can serve as input to other recent reconstruction frameworks (\eg,~\cite{sam3dteam2025sam3d3dfyimages}).~\cref{fig:first_image} shows the inference pipeline of our approach.

Across simulation experiments, we show that integrating proprioception and touch substantially improves amodal completion under severe occlusion, producing reconstructions that are both visually plausible and physically consistent in real-world scale compared to vision-only baselines. Finally, we validate real-world transfer by deploying the model on a humanoid robot with an end-effector different from those used during training, demonstrating real-world applicability and generalization across embodiments.
In summary, our contributions are:
\begin{itemize}
    \item A method to inject physical constraints into a generative reconstruction pipeline.
    \item A pose-aware, camera-aligned SDF representation and multimodal conditioning that align 2D evidence, hand pose, and tactile contacts in a shared 3D grid.
    \item Simulation and real-robot results demonstrating improved amodal completion and cross-embodiment transfer relative to vision-only baselines.
\end{itemize}

\section{Related Works}
\subsection{3D Generative Models} 

Early 3D generative approaches used GANs~\cite{goodfellow2020generative} to synthesize various 3D representations such as point clouds~\cite{huang2020pf, lin2018learning}, NeRF-style radiance fields~\cite{chan2021pi, chan2022efficient, niemeyer2021giraffe, schwarz2020graf}, or SDF~\cite{gao2022get3d} from visual inputs, but often struggled to generalize beyond the training distribution. With the emergence of Diffusion Models~\cite{sohl2015deep, ho2020denoising}, subsequent works explored representations such as point clouds~\cite{luo2021diffusion, melas2023pc2, wu2023sketch}, voxel grids~\cite{muller2023diffrf, li2023diffusion}, Triplanes~\cite{chen2023single, shue20233d, zhang2024rodinhd}, or Gaussian mixtures~\cite{zhang2024gaussiancube}, improving fidelity but frequently at high computational cost. To address efficiency, several recent methods generate shapes in a compact latent space~\cite{rombach2022high, vahdat2022lion,ren2024xcube, chen20253dtopia, zhang2024clay}, and rectified-flow formulations~\cite{albergo2022building, lipman2022flow, liu2022flow} have further accelerated sampling~\cite{xiang2024structured}. However, all these methods assume the objects are fully visible, without considering occlusions, limiting their use in real-world scenarios. In parallel, a growing body of work studies partial observability, reconstructing or generating 3D shape from incomplete inputs~\cite{chu2023diffcompletediffusionbasedgenerative3d, cui2024neusdfusion}. Recent approaches~\cite{wu2025amodal3ramodal3dreconstruction, sam3dteam2025sam3d3dfyimages, cho2025robust} explicitly model occlusions by incorporating foreground masks into the generation pipeline. However, these methods typically rely on visual cues alone and do not leverage physical interaction signals to resolve ambiguities induced by occlusion. Physical constraints have also been incorporated into human–object and hand–object interaction, using human topology, hand pose, interaction priors, pose-aligned implicit representations, or optimization-based geometric constraints for monocular reconstruction~\cite{chi2025contactawareamodalcompletionhumanobject, ye2022hand, chen2022alignsdf, petrov2023popup, easyHOI, aytekin2025follow}. Related physics-based optimization has also been introduced in category-level 3D reconstruction~\cite{yang2024physcene, ni2024phyrecon} and object-agnostic reconstruction~\cite{chen2024atlas3d, guo2024physically, li2025dso}. Separately, robotic tactile sensing has been used to improve 3D generation by fusing vision with contact observations~\cite{gao2024tactile}. In contrast, our approach integrates proprioception and multi-contact touch as 3D-space conditioning signals to guide occlusion-aware generation, enabling physically consistent completion and pose estimation.

\subsection{3D Reconstruction in Robotics}
In robotics, touch has long been used to estimate object properties~\cite{luo2025tactile}. Some works~\cite{2017-vezzani-memory,2016-jamali-active,xu2023tandem3dactivetactileexploration, zheng2024bayesianframeworkactivetactile, zheng2026bayesianactiveobjectrecognition} performed extensive tactile exploration for object recognition and pose estimation. Later methods~\cite{shahidzadeh2024actexplore, comi2024touchsdf} inferred 3D shape from multiple contacts. While effective, purely tactile approaches are inherently local and lack the global context provided by vision. Accordingly, subsequent work fused vision and touch, initially on simple shapes~\cite{wang20183d, suresh2022shapemap, tahoun2021visual, chen2023sliding} and later in more complex scenes using modern neural 3D representations~\cite{comi2025snap, comi2024touchsdf, swann2024touch}. Robotic in-hand reconstruction has also been studied with RGB-D tracking and modeling of a grasped object~\cite{manipulatorIJRR}, multi-finger contacts~\cite{smith20203d}, active strategies that select informative interactions~\cite{smith2021active}, and humanoid hands for SLAM-like reconstruction of novel objects during manipulation~\cite{suresh2024neuralfeels}. Conversely, we use touch and proprioception in the context of 3D generation under occlusion from a single RGB image, while remaining agnostic to the specific robotic hand and additionally estimating object scale and pose.
\begin{figure}[t]
	\centering
	\includegraphics[scale=0.36]{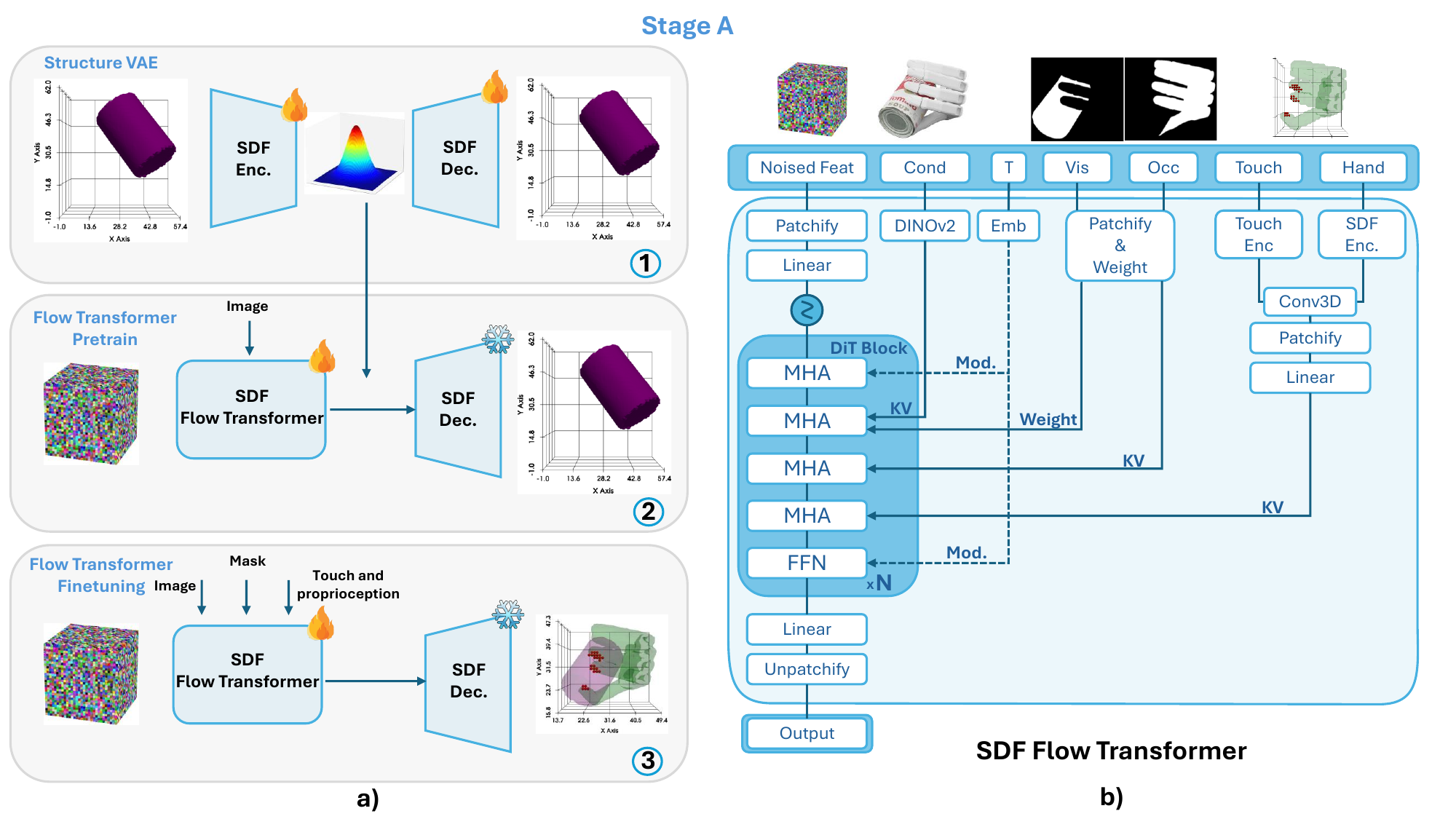}
	\caption{\textbf{Training procedure and Architecture}. \textbf{a1)}: We train a Structure-VAE autoencoder that reconstructs pose-aware object SDFs. \textbf{a2)}: Using the frozen VAE encoder, we build latent datasets and train a conditional flow transformer from scratch using pose-consistent, unoccluded object images. \textbf{a3)}: We finetune the Structure-flow on occluded manipulation scenes, conditioning on visible RGB evidence, occluder/visibility masks, hand latent, and tactile features. \textbf{b)} The architecture of the Flow Transformer. Figure design inspired by~\cite{wu2025amodal3ramodal3dreconstruction}.}
    \label{fig:architecture}
    \vskip -1,5em
\end{figure}
\section{Method}
\label{sec:method}

Given a single egocentric observation of a robot manipulating an object, our goal is to reconstruct the object's 3D geometry at metric scale and recover its appearance despite severe occlusions induced by the hand, as shown in~\cref{fig:first_image}.
For each observation we are given: (i) an RGB image $I \in \mathbb{R}^{H \times W \times 3}$, (ii) joint angles $q\in\mathbb{R}^N$ defining the robot kinematic chain (and thus the camera--hand relative pose), (iii) a hand mesh $\mesh_h$ obtained via forward kinematics, and (iv) fingertip tactile readings $u$.
Our approach focuses on \textbf{physically grounded} structure inference in a generative model.
\textbf{Stage A} predicts a coarse object shape as a continuous signed distance field (SDF), conditioning on visible RGB evidence and, during finetuning and inference, leveraging proprioception and touch to impose contact and non-interpenetration constraints.
To obtain a final textured reconstruction, we optionally pass the Stage-A output to \textbf{Stage B}, a SLat-style refinement module (Structured LAtens introduced in~\cite{xiang2024structured}) adapted to preserve pose and occlusion consistency with Stage A, or we pass the sampled voxelization to Sam3D~\cite{sam3dteam2025sam3d3dfyimages} refinement stage (further discussion in Supp. Mat.).~\cref{fig:architecture} shows the training procedure and the architecture of the Flow Transformer.\\
We next describe our dataset generation procedure (\cref{sec:data}), then detail Stage~A (\cref{subsec:stageA}), briefly describe Stage~B (\cref{subsec:stageB}), and finally outline the end-to-end inference pipeline (\cref{subsec:full_inference}). Additional details can be found in the Supp. Mat.
\subsection{Data Generation}
\label{sec:data}
We train the two stages using synthetic object renders and simulated grasp scenes.
For Stage~A and Stage~B pretraining we use isolated objects with clean renders, while the finetuning stages use grasp scenes.
All data are generated to enforce a deterministic correspondence between a rendered view and a canonical $\R^{R,R,R}$ 3D grid.
\subsubsection{Rendering Clean and Occluded Conditioning Images}
\label{sec:rendering_clean}
For each object, we render RGBA images from a predefined set of camera viewpoints sampled on a sphere around the object.
We normalize meshes to a canonical bounding box before rendering, and each rendered sample provides (i) an RGB(A) image, (ii) the camera pose, and (iii) the grid-alignment metadata used to construct the camera-aligned SDF grid.
For finetuning, we generate simulated grasp scenes by placing the object in a posed hand configuration using the network introduced by~\cite{wei2025mathcaldrograspunifiedrepresentation}.
We render an RGB image $I$ together with instance segmentation masks for the visible object and the hand/occluder, denoted $(M_o,M_h)$.
\subsubsection{Pose-Aware Metric SDF Grids}
\label{sec:sdf_rep}

To ground the 3D generation in physics, we need a continuous representation of the scene. We represent object geometry with a signed distance field (SDF) $S_o:\Omega\rightarrow\mathbb{R}$ over a cubic domain $\Omega\subset\mathbb{R}^3$ discretized on an $R^3$ grid (we use $R=64$), where the zero level set $\{\mathbf{x}:S_o(\mathbf{x})=0\}$ defines the surface. To deal with non-watertight meshes, we follow the approach of~\cite{Wang-Sig2022}.
To maintain consistency between 2D evidence and the 3D representation, we define the grid frame to be \emph{camera-aligned}: the grid $+z$ axis matches the viewing direction of the camera used to render the conditioning image, and the grid rotation is stored as metadata for each sample. The grid orientation is defined based on the views used for rendering the images, so that projecting the SDF along the grid $z$ axis is consistent with the conditioning image.

During dataset generation we further apply random in-grid similarity transforms while ensuring the surface remains inside $\Omega$.
Concretely, we sample a downscaling factor $s_{aug}\sim\mathcal{U}(0.5,1.0)$ (clipped to the maximum feasible value given the rotation and margin) and an in-plane translation $(t_x,t_y)$ while centering along $z$, obtaining the augmented SDF. The augmented SDF is obtained by evaluating the canonical field at the inverse-warped coordinates and scaling distances accordingly:
\begin{equation}
S_o^{\mathrm{aug}}(\mathbf{x})
=
s_{aug}S_o\!\left(\frac{\Rv_g^\top(\mathbf{x}-\tv)}{s_{aug}}\right),
\label{eq:sdf_aug}
\end{equation}

where $\Rv_g$ is the grid rotation and $\tv$ is the translation in grid coordinates. These poses variations are necessary to match finetuning conditions: in grasp scenes, the object and hand share the same grid, so the object’s size and location are determined by real-world scale and the hand–object relative pose, and are therefore not necessarily centered or tightly fit to the grid. 

\subsection{Stage A}\label{subsec:stageA}
\subsubsection{Structure-VAE for SDF Reconstruction}
\label{subsec:structure_vae_sdf}

We learn a compact latent space for SDF grids using a variational autoencoder.
Let $S\in\mathbb{R}^{1\times R\times R\times R}$ be an SDF grid over $\Omega=[-1,1]^3$.
A 3D convolutional encoder $E_\phi$ parameterizes a diagonal Gaussian posterior
$q_\phi(\zv\mid S)=\mathcal{N}(\boldsymbol{\mu},\mathrm{diag}(\boldsymbol{\sigma}^2))$ and produces a latent grid $\zv\in\mathbb{R}^{C\times R'\times R'\times R'}$:
\begin{equation}
(\boldsymbol{\mu},\log\boldsymbol{\sigma}^2)=E_\phi(S),
\qquad
\zv=\boldsymbol{\mu}+\boldsymbol{\sigma}\odot\boldsymbol{\epsilon},
\qquad
\boldsymbol{\epsilon}\sim\mathcal{N}(\mathbf{0},\mathbf{I}).
\label{eq:vae_reparam}
\end{equation}
A 3D convolutional decoder $D_\theta$ reconstructs the SDF grid:
\begin{equation}
\widehat{S}=D_\theta(\zv).
\label{eq:vae_decode}
\end{equation}

\subsubsection{Structure-VAE Objective}
\label{subsec:geom_loss}

We train the Structure-VAE to reconstruct SDF grids while encouraging valid signed-distance behavior. The reconstruction term is an $\ell_1$ loss over the grid,
\begin{equation}
\mathcal{L}_{\mathrm{L1}}
=
\frac{1}{|\Omega|}\sum_{\mathbf{x}\in\Omega}
\left|\widehat{S}(\mathbf{x}) - S(\mathbf{x})\right|.
\label{eq:l1_sdf}
\end{equation}
We additionally impose two geometric regularizers computed from finite-difference SDF gradients (implemented with a fixed $3{\times}3{\times}3$ convolution stencil).
To avoid boundary artifacts from padding, we evaluate gradient-based terms only on an interior mask (dropping a one-voxel rim).

\paragraph{Eikonal Regularization.}
A signed distance field satisfies $\|\nabla S(\mathbf{x})\|_2 \approx 1$ almost everywhere, thus we minimize
\begin{equation}
\mathcal{L}_{\mathrm{eik}}
=
\mathbb{E}_{\mathbf{x}}
\left(\left\|\nabla \widehat{S}(\mathbf{x})\right\|_2 - 1\right)^2,
\label{eq:eikonal_loss}
\end{equation}
where the expectation is taken over interior voxels.
\paragraph{Normal Consistency (Near-Surface).}
We encourage the predicted and ground-truth normal directions to agree in a narrow band $2\times h$ around the surface:
\begin{equation}
\mathcal{L}_{\mathrm{n}}
=
\mathbb{E}_{\mathbf{x}\,:\,|S(\mathbf{x})|<2h}
\left(1 - 
\frac{\nabla \widehat{S}(\mathbf{x})}{\|\nabla \widehat{S}(\mathbf{x})\|_2}
\cdot
\frac{\nabla S(\mathbf{x})}{\|\nabla S(\mathbf{x})\|_2}
\right),
\label{eq:normal_loss}
\end{equation}
again evaluated on interior voxels.
\paragraph{KL Divergence.}
We regularize the posterior $q_\phi(\mathbf{z}\mid S)$ toward $\mathcal{N}(\mathbf{0},\mathbf{I})$ with the standard KL term $\mathcal{L}_{\mathrm{KL}}$.\\
The overall training objective is
\begin{equation}
\mathcal{L}_{\mathrm{VAE}}
=
\lambda_{\mathrm{L1}}\,\mathcal{L}_{\mathrm{L1}}
+
\lambda_{\mathrm{eik}}\,\mathcal{L}_{\mathrm{eik}}
+
\lambda_{\mathrm{n}}\,\mathcal{L}_{\mathrm{n}}
+
\lambda_{\mathrm{KL}}\,\mathcal{L}_{\mathrm{KL}}.
\label{eq:total_geom_vae}
\end{equation}

\subsubsection{Flow Model Pre-train}
\label{sec:flow_pretraining}
We train a conditional flow-matching model in the Structure-VAE latent space to predict an object latent grid $x_0\in\mathbb{R}^{C\times R'\times R'\times R'}$ from observations.
Following standard flow matching, we sample a timestep $t\in[0,1]$, draw noise $\epsilon\sim\mathcal{N}(\mathbf{0},\mathbf{I})$, form a noisy latent $x_t$ by linear interpolation between $x_0$ and $\epsilon$ (with a small minimum noise $\sigma_{\min}$), and train a denoiser $f_\theta$ to predict the corresponding velocity field. We rescale the continuous time $t\in[0,1]$ to $\tau=\alpha t$ before the timestep embedding. During pretraining, $f_\theta$ is conditioned on a clean rendered object image $I$ that is cropped, placed to match the in-grid pose used to generate $x_0$, and then encoded using~\cite{oquab2024dinov2learningrobustvisual}.
We optimize the flow-matching objective
\begin{equation}
\mathcal{L}_{\mathrm{FM}}
=
\mathbb{E}_{x_0,t,\epsilon}\left[
\left\|
f_\theta(x_t,\alpha t;\,I)\;-\;\big((1-\sigma_{\min})\epsilon - x_0\big)
\right\|_2^2
\right],
\label{eq:flow_loss_pretrain_compact}
\end{equation}
\subsubsection{Finetuning with Occlusion, Proprioception, and Touch}
\label{sec:flow_finetune}

We finetune the flow model on simulated grasp scenes, where the object is partially occluded by the hand.
Each sample provides an RGB image $I$, a visible-object mask $M_o$, and a hand/occluder mask $M_h$.
We condition the flow model on the visible evidence $I_o = I \odot M_o$ together with the masks $(M_o,M_h)$, following~\cite{wu2025amodal3ramodal3dreconstruction} approach.

Proprioception determines the posed hand geometry in the same canonical grid as the object.
We rasterize the posed hand as an SDF grid $S_h$ and encode it with the same Structure-VAE encoder used for objects, obtaining a hand latent grid $x_{0,h}=E_\phi(S_h)$.
Touch is represented as a two-channel 3D tensor $T=[C,D]$, where $C$ is a binary contact-occupancy grid and $D$ stores, for each voxel, its distance to the nearest contact voxel.
While $C$ provides sparse surface constraints, $D$ supplies a dense, smooth field that helps learning by propagating contact information beyond the contacted voxels. To integrate touch, we encode $T$ with a lightweight 3D CNN $g_\psi$ and fuse the resulting feature volume with $x_{0,h}$ via concatenation followed by a $1\times 1\times 1$ projection:
\begin{equation}
\widetilde{x}_{0,h}
=
\mathrm{Conv}_{1\times 1\times 1}\!\left(\big[x_{0,h},\ g_\psi(T)\big]\right),
\label{eq:fuse_hand_touch}
\end{equation}
so that $\widetilde{x}_{0,h}$ reduces to $x_{0,h}$ when touch is disabled.
The fused volume is patchified into tokens and provided to the denoiser via an additional cross-attention stream, alongside the image-based conditioning.

\paragraph{Physics-aware finetuning losses.}
In addition to the flow-matching regression loss $\mathcal{L}_{\mathrm{FM}}$, we impose auxiliary physical losses computed on decoded SDFs.
Given the denoiser prediction, we form a clean latent estimate and decode it with the \emph{frozen} Structure-VAE decoder to obtain the predicted object SDF $\widehat{S}_o$; we decode the hand latent $x_{0,h}$ once to obtain the hand SDF $S_h$.
Because predictions at large diffusion timesteps are highly noisy, we downweight physics terms as a function of time using
\begin{equation}
w(t)=(1-t)^2,
\label{eq:time_weight}
\end{equation}
and compute a weighted batch mean.\\
We penalize object mass inside the hand volume using a smooth interior saturation to avoid unstable gradients from deep penetration.
Let $\tau$ be a saturation threshold (we use $\tau=0.1$) and define
\begin{equation}
\psi_\tau(s)=\tau\,\tanh\!\Big(\mathrm{ReLU}(s)/\tau\Big).
\label{eq:saturation}
\end{equation}
We compute the saturated interior field for the object and a binary hand-volume mask:
\(
A_o(\x)=\psi_\tau(-\widehat{S}_o(\x))
\)
and
\(
M_h(\x)=\mathbf{1}\!\left[\psi_\tau(-S_h(\x))>0\right].
\)
The non-interpenetration loss is the mean object interior mass inside the hand volume,
\begin{equation}
\mathcal{L}_{\mathrm{NI}}
=
\frac{1}{B}\sum_{b=1}^{B}
\frac{\sum_{\x} A_{o,b}(\x)\,M_{h,b}(\x)}
{\max\!\left(1,\sum_{\x} M_{h,b}(\x)\right)},
\label{eq:ni_loss}
\end{equation}
and we apply the time-weighted reduction described above.\\
Let $C(\x)\in\{0,1\}$ denote the binary contact grid (first channel of $T$).
We encourage the reconstructed surface to pass through contact locations by penalizing the SDF magnitude at contact voxels:
\begin{equation}
\mathcal{L}_{\mathrm{C}}
=
\frac{1}{B}\sum_{b=1}^{B}
\frac{\sum_{\x} C_b(\x)\,|\widehat{S}_{o,b}(\x)|}
{\max\!\left(1,\sum_{\x} C_b(\x)\right)},
\label{eq:contact_loss_method}
\end{equation}
again using the time-weighted batch reduction.\\
The overall finetuning objective is
\begin{equation}
\mathcal{L}
=
\mathcal{L}_{\mathrm{FM}}
+
\lambda_{\mathrm{NI}}\,\mathcal{L}_{\mathrm{NI}}
+
\lambda_{\mathrm{C}}\,\mathcal{L}_{\mathrm{C}},
\label{eq:flow_loss_total}
\end{equation}
where $\lambda_{\mathrm{NI}}$ is warmed up during training and $\lambda_{\mathrm{C}}$ is fixed.

\subsection{Stage B}
\label{subsec:stageB}

Stage~B is included primarily as an integration step to demonstrate that the metric-scale, physically consistent structure predicted by Stage~A can be plugged into a standard refinement/texturing pipeline. We adopt a SLat-style representation following~\cite{xiang2024structured}, with minimal adaptations for manipulation.
We derive posed sparse voxel features from the Stage~A prediction in the same camera-aligned canonical grid.
SLAT decoding is performed in this pose-aligned frame and mapped to the world/camera frame using the known similarity transform from the pose/normalization metadata before rendering supervision.
To predict the SLAT latent from an observation, we condition on the visible object evidence $I_o=I\odot M_o$ together with the hand/occluder mask $M_h$ (as in~\cite{wu2025amodal3ramodal3dreconstruction}), and train with the same flow-matching objective as Stage~A (\cref{eq:flow_loss_pretrain_compact}).
\subsection{Physical guidance at inference}
\label{subsec:full_inference}

At inference time, we sample the Stage~A latent with an explicit Euler solver driven by the learned velocity field $v_\vartheta(\cdot)$ under conditioning \emph{cond} (occlusion-aware visual cues and, when available, hand/touch signals).
To enforce physical plausibility at inference time, we introduce an additive control term $\theta_k$ in the denoising process:
\begin{equation}
x_{k+1}
=
x_k
-
\Delta t_k\Big(
v_\vartheta(x_k,t_k;\mathrm{cond})+\theta_k
\Big).
\label{eq:oc_euler}
\end{equation}
When guidance is enabled, we decode the current latent estimate with the frozen Structure-VAE decoder to obtain $\widehat{S}_o$ (and decode the fixed hand latent once to obtain $S_h$), and define the physics energy
\begin{equation}
E(x_k)
=
\lambda_{\mathrm{NI}}\,\mathcal{L}_{\mathrm{NI}}\!\big(\widehat{S}_o,S_h\big)
+
\lambda_{\mathrm{C}}\,\mathcal{L}_{\mathrm{C}}\!\big(\widehat{S}_o,T\big).
\label{eq:oc_energy}
\end{equation}
We compute $g_k=\nabla_{x_k}E(x_k)$ and update the control term with an exponential moving average,
\begin{equation}
\theta_{k+1} \;=\; \beta\,\theta_k \;+\; \eta\,g_k,
\label{eq:oc_theta_update}
\end{equation}
where $\beta\in(0,1)$ is a decay factor and $\eta>0$ controls guidance strength.
We apply standard stabilization heuristics (gradient normalization, a trust-region on $\|\theta_k\|$ relative to $\|v_\vartheta\|$, and optional projection to prevent increasing the contact energy) inspired by~\cite{wang2024training}.

\section{Experiments}
\subsection{Dataset and Metrics.}
For VAE training and Flow Transformer pretraining, we use 3D-FUTURE~\cite{fu20213d}, HSSD~\cite{khanna2024habitat}, ABO~\cite{collins2022abo}, and a subset of ObjaverseXL~\cite{deitke2023objaverse}($100k$ objects). For the Flow Transformer finetuning, we use Google Scanned Objects~\cite{downs2022google}. We
test on a subset of the YCB dataset~\cite{7254318} consisting of 36 objects.
Grasps are generated as in~\cref{sec:rendering_clean}, yielding $22{,}680$ test images. 
We stratify results by occlusion level using $K{=}5$ equal-width bins over $x \in [0,1]$ (for simplicity, we refer to them as $B_{k}$) with a non-even distribution of samples.
We evaluate reconstruction using metrics commonly used in the literature: Chamfer Distance (\textbf{CD}), Normal consistency (\textbf{NC}), F-score with a $0.02$ threshold (\textbf{F@0.02}), Voxel Intersection over Union (\textbf{Voxel-IoU}), and Earth Mover's Distance (\textbf{EMD}). For pose estimation, we use 3D Intersection over Union (\textbf{3D IoU}), \textbf{ICP-Rot} as defined in~\cite{sam3dteam2025sam3d3dfyimages}, Average Distance with Symmetry (\textbf{ADD-S}) from~\cite{xiang2018posecnnconvolutionalneuralnetwork}, and at a threshold of $0.1$ \textbf{ADD-S@0.1}. Only valid output meshes are considered. More information and comparison with~\cite{aytekin2025follow,easyHOI} can be found in the Supp. Mat.

\begin{table}[ht]
    \centering
    \caption{3D reconstruction results in simulation.}
    \label{tab:3drecon}
    \resizebox{0.7\textwidth}{!}{
    \begin{tabular}{c|c|c|c|c|c|c}
        \hline
        &  & \textbf{CD} $\downarrow$ & \textbf{NC} $\uparrow$ & \textbf{F@0.02} $\uparrow$ & \textbf{Voxel IoU} $\uparrow$ & \textbf{EMD} $\downarrow$ \\
        \hline

        \multirow{4}{*}{\textbf{Amodal3R}~\cite{wu2025amodal3ramodal3dreconstruction}} & \bm{$B_{1}$} & 0.126 & 0.704 & 0.178 & $0.386$&  $0.287$ \\
                                        & \bm{$B_{2}$} & 0.150 & 0.674 & 0.188 & $0.331$  & $0.291$ \\
                                        & \bm{$B_{3}$} & 0.212 & 0.644 & 0.156 & $0.302$ &$0.333$ \\
                                        & \bm{$B_{4}$} & 0.263 & 0.624 & 0.120 & $0.295$ &$0.374$  \\
                                        & \bm{$B_{5}$} & 0.439 & 0.602 & 0.078 & $0.262$ &$0.406$  \\
                                        \cline{2-7}
                                        & \textbf{All} & $0.188$ & $0.669$ & $0.162$ & $0.339$ & $0.314$ \\
        \hline

        \multirow{4}{*}{\textbf{Sam3D}~\cite{sam3dteam2025sam3d3dfyimages}}  & \bm{$B_{1}$} & \bm{$0.020$} & $0.837$ & \bm{$0.265$} & $0.558$ & $0.172$ \\
                                        & \bm{$B_{2}$} & $0.025$ & $0.811$ & \bm{$0.248$} &  $0.514$ & $0.184$ \\
                                        & \bm{$B_{3}$} & $0.034$ & $0.081$ & \bm{$0.198$} &  $0.490$ & $0.201$ \\
                                        & \bm{$B_{4}$} & $0.057$  & $0.775$ & \bm{$0.137$} &  $0.433$ & $0.239$ \\
                                        & \bm{$B_{5}$} & $0.153$  & $0.732$ & $0.085$ & $0.337$ & $0.342$ \\
                                        \cline{2-7}
                                        & \textbf{All} & $0.039$ & $0.803$ & \bm{$0.228$} & $0.504$ & $0.202$\\
        \hline

        \multirow{4}{*}{\textbf{Ours}}   & \bm{$B_{1}$} & $0.021$ & \bm{$0.868$}  & $0.215$ & \bm{$0.616$}  & \bm{$0.158$} \\
                                        & \bm{$B_{2}$} & \bm{$0.022$} & \bm{$0.844$} & $0.208$ & \bm{$0.585$}  & \bm{$0.172$} \\
                                        & \bm{$B_{3}$} & \bm{$0.027$} & \bm{$0.834$} & $0.180$ & \bm{$0.576$} & \bm{$0.188$} \\
                                        & \bm{$B_{4}$} & \bm{$0.055$} & \bm{$0.805$} & \bm{$0.137$} & \bm{$0.526$} & \bm{$0.229$} \\
                                        & \bm{$B_{5}$} & \bm{$0.109$} & \bm{$0.803$} & \bm{$0.101$} & \bm{$0.532$} & \bm{$0.276$} \\
                                        \cline{2-7}
                                        & \textbf{All} & \bm{$0.033$} & \bm{$0.844$} & $0.189$ & \bm{$0.586$} & \bm{$0.184$}\\
        \hline
    \end{tabular}}
\end{table}

\begin{figure}[!t]
  \centering

  \includegraphics[scale=0.35]{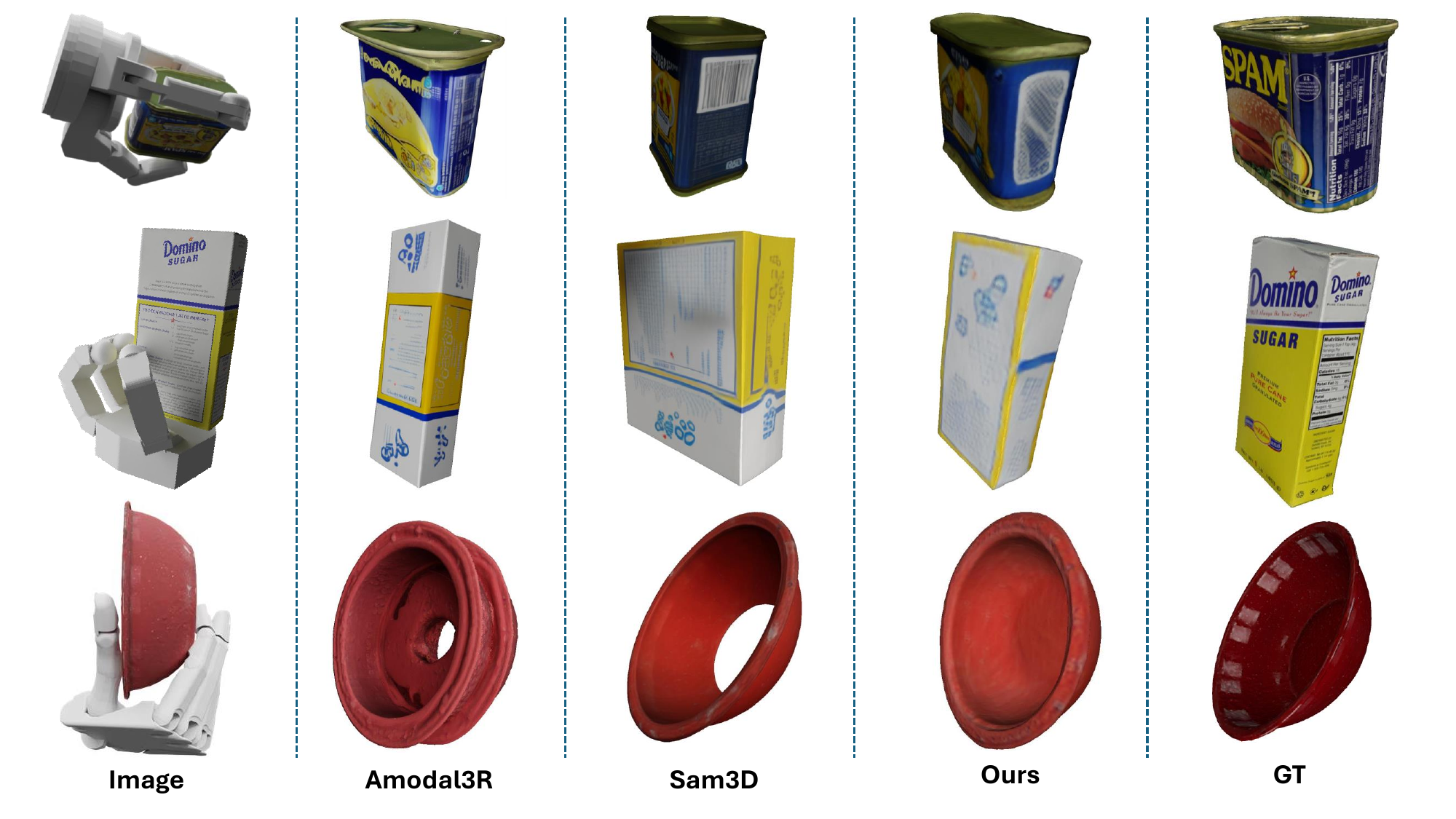}
  \caption{\textbf{Qualitative results in simulation}. Vision-only baselines often exhibit artifacts (\eg, holes or inconsistent relative dimensions) under occlusion. By leveraging contact cues, our method produces more physically plausible reconstructions.}
  \label{fig:qualitative_results}
\end{figure}
\subsection{Simulation Results}
\subsubsection{3D reconstruction} We compare our method against two representative approaches for 3D reconstruction under occlusion: Amodal3R~\cite{wu2025amodal3ramodal3dreconstruction} and SAM3D~\cite{sam3dteam2025sam3d3dfyimages}. Amodal3R is a natural baseline for our setting, as it builds upon the same TRELLIS-style architecture, but relies on \emph{vision-only} conditioning. SAM3D serves as a strong contemporary baseline for single-image 3D reconstruction. Both baselines are trained with larger datasets than ours, and SAM3D uses a larger model capacity.
\cref{tab:3drecon} reports 3D reconstruction performance across occlusion bins and on the full evaluation set. Overall, our method and SAM3D outperform Amodal3R across metrics, highlighting the limitations of vision-only conditioning in heavily occluded in-hand scenarios. Compared to SAM3D, incorporating proprioception, touch, and physics-based constraints improves performance on all metrics, except in the least-occluded regime for $\mathbf{F@0.02}$.
This suggests that when occlusion is mild, a strong vision-only model can match or slightly exceed our reconstruction quality, whereas its performance degrades more substantially as occlusion increases. Qualitative comparisons are shown in~\cref{fig:qualitative_results}. 
\begin{table}[ht]
    \centering
    \caption{Pose estimation results in simulation.}
    \label{tab:pose}
    \resizebox{0.7\textwidth}{!}{
    \begin{tabular}{c|c|c|c|c|c}
        \hline
        &  & \textbf{3D IoU} $\uparrow$ & \textbf{ICP-Rot} $\downarrow$ & \textbf{ADD-S} $\downarrow$ & \textbf{ADD-S@0.1} $\uparrow$   \\
        \hline

        \multirow{4}{*}{\textbf{Sam3D}~\cite{sam3dteam2025sam3d3dfyimages}}  & \bm{$B_{1}$} & $0.453$ & $15.848$ & $0.080$ & $0.776$  \\
                                        & \bm{$B_{2}$} & $0.460$ & $18.593$ & $0.078$ &  $0.778$  \\
                                        & \bm{$B_{3}$} & $0.413$ & $19.610$ & $0.091$ &  $0.656$ \\
                                        & \bm{$B_{4}$} & $0.325$ & $22.672$  & $0.11$ &  $0.370$  \\
                                        & \bm{$B_{5}$} & $0.150$ & $35.013$ & $0.19$ &   $0.099$ \\
                                        \cline{2-6}
                                        & \textbf{All} & $0.406$ & $18.875$ & $0.095$ & $0.658$ \\
        \hline

        \multirow{4}{*}{\textbf{Ours}}   & \bm{$B_{1}$} & \bm{$0.595$} & \bm{$7.067$} & \bm{$0.062$} &  \bm{$0.864$} \\
        & \bm{$B_{2}$} & \bm{$0.551$} & \bm{$9.922$} & \bm{$0.071$} & \bm{$0.811$}   \\
                                        & \bm{$B_{3}$} & \bm{$0.512$} & \bm{$12.434$} & \bm{$0.082$} & \bm{$0.721$} \\
                                        & \bm{$B_{4}$} & \bm{$0.456$} & \bm{$16.939$} & \bm{$0.102$} &  \bm{$0.553$}  \\
                                        & \bm{$B_{5}$} & \bm{$0.356$} & \bm{$18.199$} & \bm{$0.143$} &  \bm{$0.258$}  \\
                                        \cline{2-6}
                                        & \textbf{All} & \bm{$0.53$} & \bm{$10$} & \bm{$0.07$} & \bm{$0.73$} \\
        \hline
    \end{tabular}}
\end{table}
\subsubsection{Pose estimation} For pose estimation, we compare against SAM3D~\cite{sam3dteam2025sam3d3dfyimages}. 
Because SAM3D requires 3D information, we compare against the best case in which point maps are computed from ground-truth depth; this is because in noisy conditions, pose estimates of SAM3D are significantly less accurate.
In contrast, our method does not require depth, as it leverages robot encoders to recover the camera--hand configuration, and uses touch as an additional constraint.
\cref{tab:pose} summarizes the results. Across all reported metrics, incorporating physical constraints during generation improves pose estimation accuracy. This is expected: proprioception constrains the hand pose directly, and multi-contact touch further reduces the set of object poses consistent with the observation by imposing contact and non-interpenetration feasibility.
\subsection{Real-world Results}
\begin{figure}[!t]
  \centering

  \includegraphics[scale=0.35]{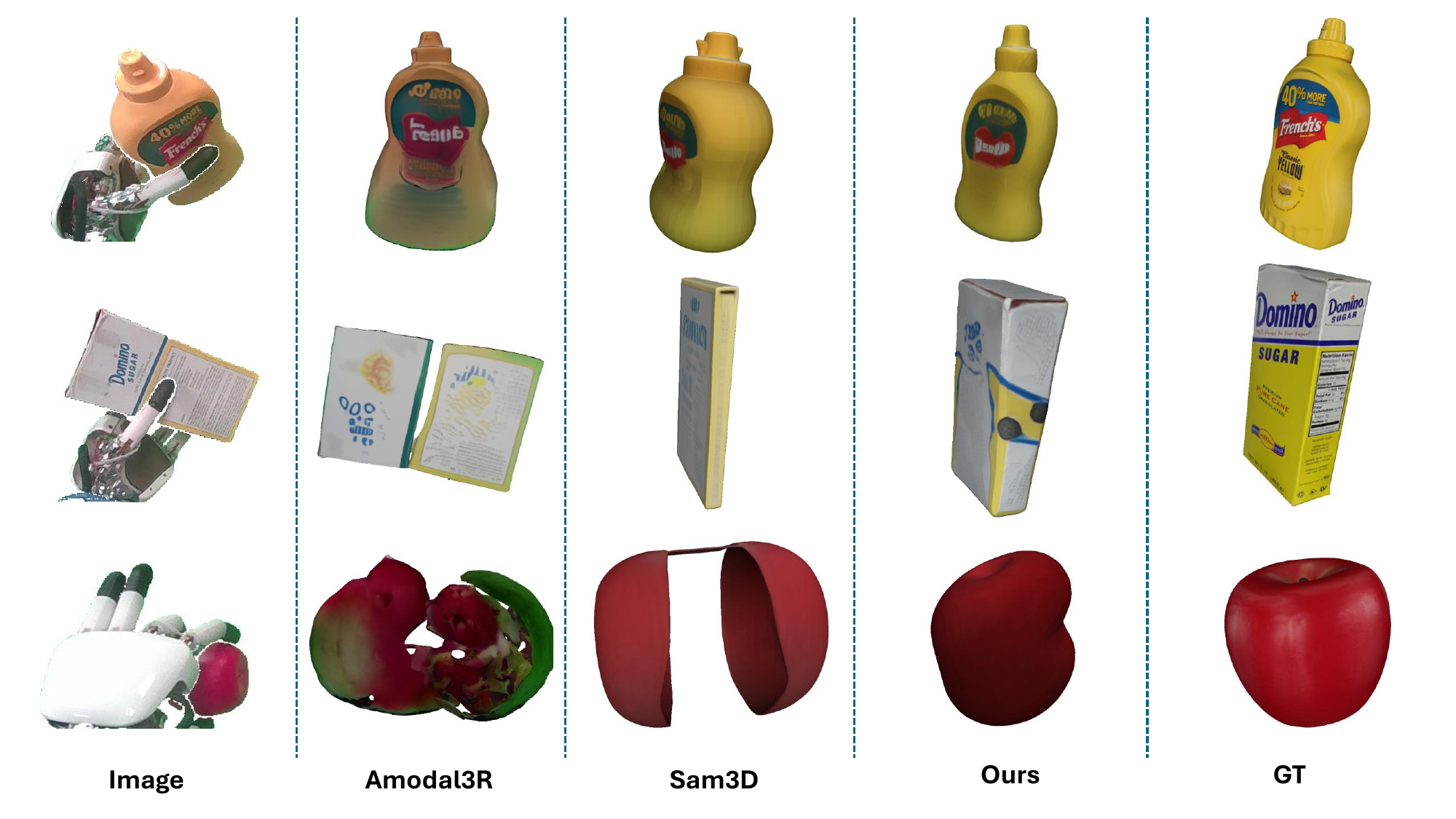}
  \caption{\textbf{Qualitative results on real-world data}. Contact cues improve physical plausibility under occlusion. Artifacts can arise when camera–hand calibration/forward kinematics are inaccurate, leading to hand–object misalignment (third row).}
  \label{fig:real_qualitative}
\end{figure}

We evaluate real-world transfer by collecting data on a humanoid robot (\cref{fig:first_image}) equipped with an end-effector different from the one used in simulation and instrumented with fingertip tactile sensors. We perform experiments on real counterparts of a subset of the objects used in simulation. For each trial, we acquire an egocentric RGB image and use calibrated camera--hand extrinsics (together with forward kinematics) to place the posed end-effector into the shared sparse-structure grid, alongside the tactile measurements.~\cref{tab:real} reports quantitative results on reconstruction. Overall, SAM3D exhibits comparable performance under noisy real-world imagery w.r.t. the simulated scenario, while Amodal3R significantly suffers in the noisy scenario. Our method attains slightly lower scores in this setting, which we attribute primarily to encoder and calibration noise affecting the hand pose and touch alignment. Qualitative examples are shown in~\cref{fig:real_qualitative}. 
\begin{table}[ht]
    \centering
    \caption{3D reconstruction results in the real-world scenario.}
    \label{tab:real}
    \resizebox{0.7\textwidth}{!}{
    \begin{tabular}{c|c|c|c|c|c}
        \hline
        &  \textbf{CD} $\downarrow$ & \textbf{NC} $\uparrow$ & \textbf{F@0.02} $\uparrow$ & \textbf{Voxel IoU} $\uparrow$ & \textbf{EMD} $\downarrow$ \\
        \hline

        \textbf{Amodal3R}~\cite{wu2025amodal3ramodal3dreconstruction}  & $0.252$ & $0.501$ & $0.083$ & $0.352$ & $0.397$\\
        \hline

        \textbf{Sam3D}~\cite{sam3dteam2025sam3d3dfyimages}  & $0.037$ & $0.888$ & \bm{$0.225$} & $0.510$ & \bm{$0.185$}\\
        \hline

        \textbf{Ours} & \bm{$0.035$} & \bm{$0.903$} & $0.186$ & \bm{$0.648$} & $0.190$\\
        \hline
    \end{tabular}}
\end{table}

\subsection{Ablation Study}
\label{subsec:ablation}
\begin{table}[ht]
    \centering
    \caption{Ablation study for 3D reconstruction in simulation. We report aggregate results over all object categories.}
    \label{tab:ablation}
    \resizebox{\columnwidth}{!}{%
    \begin{tabular}{l|c|c|c|c|c}
        \hline
        \textbf{Setting}
        & \textbf{CD} $\downarrow$
        & \textbf{NC} $\uparrow$
        & \textbf{F@0.02} $\uparrow$
        & \textbf{Voxel IoU} $\uparrow$
        & \textbf{EMD} $\downarrow$ \\
        \hline

        \multicolumn{6}{c}{\textbf{Sensing modalities}} \\
        \hline
        Vision-Only
        & $0.142$ & $0.696$ & $0.116$ & $0.357$ & $0.296$ \\
        No-Touch
        & $0.085$ & $0.760$ & $0.126$ & $0.430$ & $0.264$ \\
        \hline

        \multicolumn{6}{c}{\textbf{Input noise}} \\
        \hline
        Mask noise, level 1
        & $0.033$ & $0.835$ & $0.180$ & $0.572$ & $0.187$ \\
        Mask noise, level 2
        & $0.035$ & $0.806$ & $0.178$ & $0.537$ & $0.199$ \\
        Kinematic noise, level 1
        & $0.037$ & $0.809$ & $0.177$ & $0.541$ & $0.190$ \\
        Kinematic noise, level 2
        & $0.042$ & $0.798$ & $0.171$ & $0.501$ & $0.212$ \\
        Tactile noise, $3$\,mm
        & $0.033$ & $0.845$ & $0.188$ & $0.575$ & $0.187$ \\
        Tactile noise, $5$\,mm
        & $0.036$ & $0.854$ & $0.186$ & $0.563$ & $0.183$ \\
        \hline

        \multicolumn{6}{c}{\textbf{Model components}} \\
        \hline
        Inference without guidance
        & $0.036$ & $0.811$ & $0.183$ & $0.530$ & $0.194$ \\
        Training without physics losses
        & $0.052$ & $0.774$ & $0.149$ & $0.487$ & $0.210$ \\
        Binary contact alone
        & $0.046$ & $0.798$ & $0.153$ & $0.505$ & $0.198$ \\
        Distance contact alone
        & $0.037$ & $0.816$ & $0.181$ & $0.532$ & $0.191$ \\
        \hline

        \textbf{Ours}
        & \bm{$0.033$} & \bm{$0.844$} & \bm{$0.189$} & \bm{$0.586$} & \bm{$0.184$} \\
        \hline
    \end{tabular}%
    }
\end{table}

We perform ablations to quantify the contribution of sensing modalities, robustness to noisy inputs, and the impact of the main training and inference components. 
Results are summarized in Table~\ref{tab:ablation}, with aggregate metrics over all object categories. Additional qualitative examples and experiments on an unseen end-effector~\cite{9578786} are provided in the Supp. Mat. 

\textbf{Sensing ablation.}
We isolate each sensing modality by training reduced-input variants of Stage~A. The \emph{Vision-Only} variant uses only visual observations, making it comparable in information content to Amodal3R~\cite{wu2025amodal3ramodal3dreconstruction}, up to differences in pose-consistent conditioning and mask distribution. It performs substantially worse than the full model, confirming that vision alone is insufficient under severe hand-object occlusion. Adding proprioception, i.e., vision plus posed end-effector geometry but no touch, clearly improves over Vision-Only by providing global geometric constraints. However, the \emph{No-Touch} variant remains below the full model, showing that tactile cues provide complementary local surface evidence at contacts. 

\textbf{Robustness to noisy observations.}
We evaluate robustness by perturbing the inputs at test time. For tactile sensing, we add random spatial noise of up to $3$\,mm and $5$\,mm to the contact locations. A $3$\,mm perturbation has negligible impact, while $5$\,mm noise causes a small degradation in CD and voxel IoU. This behavior is consistent with the discretization of our representation: the hand--object scene spans tens of centimeters and is represented on a fixed $R^3$ grid, so small perturbations are partially absorbed by voxel quantization.

We further perturb the visual masks and kinematics to identify the main sources of error in the real-world setting. For masks, levels 1 and 2 correspond to erosion/dilation and jitter with maximum perturbations of $2$ and $5$ pixels, respectively. For kinematics, levels 1 and 2 correspond to maximum hand-eye calibration errors of $5$\,mm and $1$\,cm. The results show that the model is more sensitive to kinematic noise than to mask noise, which explains part of the performance drop observed in the real-world experiments.

\textbf{Training and inference components.}
We also ablate inference guidance and the physics losses used during training. Removing inference guidance produces only a limited degradation in 3D reconstruction, suggesting that the learned model already captures most of the reconstruction signal. In contrast, removing the physics losses causes a substantial drop across all metrics, showing that they are a crucial component of the training recipe. We then isolate the contribution of the tactile representation by using either binary contact or distance-based contact. Distance-based contact explains most of the tactile improvement, while combining it with binary contact yields the best overall performance.

\section{Limitations}
One limitation of our method is that it relies on accurate pose and calibration: camera--hand extrinsics, forward kinematics, and reliable occluder/visibility masks are required to align RGB images with the 3D grid, and errors in these signals can bias the reconstruction. In addition, operating on a fixed-resolution grid ($64^3$) limits geometric fidelity, particularly for thin structures and concavities, which may be lost in Stage~A and only partially recovered downstream. This effect is amplified in manipulation scenes because the object is reconstructed in a shared grid with the end-effector, reducing the effective resolution available for the object. Finally, we notice that an object-centric frame could simplify fusing multiple views in dynamic scenes compared with the proposed camera-aligned frame. Further studies could investigate this potential issue.

\section{Conclusion}
\label{sec:conclusion}
We presented a multimodal method for metric-scale amodal object reconstruction under severe hand occlusion, and a practical way to inject physics into generative sparse-structure inference.
Our approach fuses egocentric RGB evidence with proprioceptive hand geometry and multi-contact touch, representing shape as a pose-aware, camera-aligned SDF learned with a Structure-VAE.
A conditional flow-matching model is pretrained on vision-only data and finetuned on occluded manipulation scenes, while contact-consistency and non-interpenetration objectives are used both as training losses and as decoder-guided sampling terms at inference time, steering generation toward physically feasible shapes.
Experiments in simulation show that incorporating proprioception and touch improves occlusion-robust geometry reconstruction over vision-only baselines and yields physically plausible shapes at the correct real-world scale.
We further demonstrate transfer to a real humanoid robot with an unseen end-effector, highlighting the benefit of expressing interaction cues in a shared 3D frame.

Future work includes scaling to larger, more diverse object–interaction datasets, developing more efficient conditioning than multi-stream cross-attention, and integrating contact cues more tightly in a single-stage pipeline. Beyond geometry, touch could also be used to infer physical properties such as friction or compliance, enabling reconstructions that better predict interaction dynamics.

\textbf{Acknowledgements.} This work was supported by the National Institute for Insurance against Accidents at Work (INAIL) project ergoCub-core. We thank Lorenzo Paiola for the inspiring and valuable discussions.

\bibliographystyle{splncs04}
\bibliography{main}

@String(CVPR  = {IEEE Conf. Comput. Vis. Pattern Recog.})

@String(ECCV  = {Eur. Conf. Comput. Vis.})

@String(AAAI  = {AAAI})

@String(TOG   = {ACM Trans. Graph.})

@String(CVPR  = {CVPR})

@String(ECCV  = {ECCV})

@String(TOG   = {ACM TOG})

@article{xiang2024structured,
    title   = {Structured 3D Latents for Scalable and Versatile 3D Generation},
    author  = {Xiang, Jianfeng and Lv, Zelong and Xu, Sicheng and Deng, Yu and Wang, Ruicheng and Zhang, Bowen and Chen, Dong and Tong, Xin and Yang, Jiaolong},
    journal = {arXiv preprint arXiv:2412.01506},
    year    = {2024}
}

@misc{wu2025amodal3ramodal3dreconstruction,
      title={Amodal3R: Amodal 3D Reconstruction from Occluded 2D Images}, 
      author={Tianhao Wu and Chuanxia Zheng and Frank Guan and Andrea Vedaldi and Tat-Jen Cham},
      year={2025},
      eprint={2503.13439},
      archivePrefix={arXiv},
      primaryClass={cs.CV},
      url={https://arxiv.org/abs/2503.13439}, 
}

@article{suresh2024neuralfeels,
  title={NeuralFeels with neural fields: Visuotactile perception for in-hand manipulation},
  author={Suresh, Sudharshan and Qi, Haozhi and Wu, Tingfan and Fan, Taosha and Pineda, Luis and Lambeta, Mike and Malik, Jitendra and Kalakrishnan, Mrinal and Calandra, Roberto and Kaess, Michael and others},
  journal={Science Robotics},
  volume={9},
  number={96},
  pages={eadl0628},
  year={2024},
  publisher={American Association for the Advancement of Science}
}

@article{gao2024tactile,
  title={Tactile DreamFusion: Exploiting tactile sensing for 3D generation},
  author={Gao, Ruihan and Deng, Kangle and Yang, Gengshan and Yuan, Wenzhen and Zhu, Jun-Yan},
  journal={Advances in Neural Information Processing Systems},
  volume={37},
  pages={29839--29863},
  year={2024}
}

@inproceedings{comi2025snap,
  title={Snap-it, tap-it, splat-it: Tactile-informed 3d gaussian splatting for reconstructing challenging surfaces},
  author={Comi, Mauro and Tonioni, Alessio and Tremblay, Jonathan and Yang, Max and Blukis, Valts and Lin, Yijiong and Lepora, Nathan F and Aitchison, Laurence},
  booktitle={2025 International Conference on 3D Vision (3DV)},
  pages={1134--1143},
  year={2025},
  organization={IEEE}
}

@inproceedings{shahidzadeh2024actexplore,
  title={Actexplore: Active tactile exploration on unknown objects},
  author={Shahidzadeh, Amir-Hossein and Yoo, Seong Jong and Mantripragada, Pavan and Singh, Chahat Deep and Ferm{\"u}ller, Cornelia and Aloimonos, Yiannis},
  booktitle={2024 IEEE International Conference on Robotics and Automation (ICRA)},
  pages={3411--3418},
  year={2024},
  organization={IEEE}
}

@article{comi2024touchsdf,
  title={Touchsdf: A deepsdf approach for 3d shape reconstruction using vision-based tactile sensing},
  author={Comi, Mauro and Lin, Yijiong and Church, Alex and Tonioni, Alessio and Aitchison, Laurence and Lepora, Nathan F},
  journal={IEEE Robotics and Automation Letters},
  volume={9},
  number={6},
  pages={5719--5726},
  year={2024},
  publisher={IEEE}
}

@misc{chi2025contactawareamodalcompletionhumanobject,
      title={Contact-Aware Amodal Completion for Human-Object Interaction via Multi-Regional Inpainting}, 
      author={Seunggeun Chi and Enna Sachdeva and Pin-Hao Huang and Kwonjoon Lee},
      year={2025},
      eprint={2508.00427},
      archivePrefix={arXiv},
      primaryClass={cs.CV},
      url={https://arxiv.org/abs/2508.00427}, 
}

@article{sam3dteam2025sam3d3dfyimages,
      title={SAM 3D: 3Dfy Anything in Images}, 
      author={SAM 3D Team and Xingyu Chen and Fu-Jen Chu and Pierre Gleize and Kevin J Liang and Alexander Sax and Hao Tang and Weiyao Wang and Michelle Guo and Thibaut Hardin and Xiang Li and Aohan Lin and Jiawei Liu and Ziqi Ma and Anushka Sagar and Bowen Song and Xiaodong Wang and Jianing Yang and Bowen Zhang and Piotr Dollár and Georgia Gkioxari and Matt Feiszli and Jitendra Malik},
      year={2025},
      eprint={2511.16624},
      archivePrefix={arXiv},
      primaryClass={cs.CV},
      url={https://arxiv.org/abs/2511.16624}, 
}

@article{chu2023diffcompletediffusionbasedgenerative3d,
  title={Diffcomplete: Diffusion-based generative 3d shape completion},
  author={Chu, Ruihang and Xie, Enze and Mo, Shentong and Li, Zhenguo and Nie{\ss}ner, Matthias and Fu, Chi-Wing and Jia, Jiaya},
  journal={Advances in Neural Information Processing Systems},
  year={2023}
}

@inproceedings{cui2024neusdfusion,
  title={Neusdfusion: A spatial-aware generative model for 3d shape completion, reconstruction, and generation},
  author={Cui, Ruikai and Liu, Weizhe and Sun, Weixuan and Wang, Senbo and Shang, Taizhang and Li, Yang and Song, Xibin and Yan, Han and Wu, Zhennan and Chen, Shenzhou and others},
  booktitle={European Conference on Computer Vision},
  pages={1--18},
  year={2024},
  organization={Springer}
}

@inproceedings{cho2025robust,
  title={Robust 3d shape reconstruction in zero-shot from a single image in the wild},
  author={Cho, Junhyeong and Youwang, Kim and Yang, Hunmin and Oh, Tae-Hyun},
  booktitle={Proceedings of the Computer Vision and Pattern Recognition Conference},
  pages={22786--22798},
  year={2025}
}

@article{goodfellow2020generative,
  title={Generative adversarial networks},
  author={Goodfellow, Ian and Pouget-Abadie, Jean and Mirza, Mehdi and Xu, Bing and Warde-Farley, David and Ozair, Sherjil and Courville, Aaron and Bengio, Yoshua},
  journal={Communications of the ACM},
  volume={63},
  number={11},
  pages={139--144},
  year={2020},
  publisher={ACM New York, NY, USA}
}

@inproceedings{huang2020pf,
  title={Pf-net: Point fractal network for 3d point cloud completion},
  author={Huang, Zitian and Yu, Yikuan and Xu, Jiawen and Ni, Feng and Le, Xinyi},
  booktitle={Proceedings of the IEEE/CVF conference on computer vision and pattern recognition},
  pages={7662--7670},
  year={2020}
}

@inproceedings{chan2021pi,
  title={pi-gan: Periodic implicit generative adversarial networks for 3d-aware image synthesis},
  author={Chan, Eric R and Monteiro, Marco and Kellnhofer, Petr and Wu, Jiajun and Wetzstein, Gordon},
  booktitle={Proceedings of the IEEE/CVF conference on computer vision and pattern recognition},
  pages={5799--5809},
  year={2021}
}

@article{gao2022get3d,
  title={Get3d: A generative model of high quality 3d textured shapes learned from images},
  author={Gao, Jun and Shen, Tianchang and Wang, Zian and Chen, Wenzheng and Yin, Kangxue and Li, Daiqing and Litany, Or and Gojcic, Zan and Fidler, Sanja},
  journal={Advances in neural information processing systems},
  volume={35},
  pages={31841--31854},
  year={2022}
}

@article{ho2020denoising,
  title={Denoising diffusion probabilistic models},
  author={Ho, Jonathan and Jain, Ajay and Abbeel, Pieter},
  journal={Advances in neural information processing systems},
  volume={33},
  pages={6840--6851},
  year={2020}
}

@inproceedings{sohl2015deep,
  title={Deep unsupervised learning using nonequilibrium thermodynamics},
  author={Sohl-Dickstein, Jascha and Weiss, Eric and Maheswaranathan, Niru and Ganguli, Surya},
  booktitle={International conference on machine learning},
  pages={2256--2265},
  year={2015},
  organization={pmlr}
}

@inproceedings{luo2021diffusion,
  title={Diffusion probabilistic models for 3d point cloud generation},
  author={Luo, Shitong and Hu, Wei},
  booktitle={Proceedings of the IEEE/CVF conference on computer vision and pattern recognition},
  pages={2837--2845},
  year={2021}
}

@inproceedings{muller2023diffrf,
  title={Diffrf: Rendering-guided 3d radiance field diffusion},
  author={M{\"u}ller, Norman and Siddiqui, Yawar and Porzi, Lorenzo and Bulo, Samuel Rota and Kontschieder, Peter and Nie{\ss}ner, Matthias},
  booktitle={Proceedings of the IEEE/CVF Conference on Computer Vision and Pattern Recognition},
  pages={4328--4338},
  year={2023}
}

@article{zhang2024gaussiancube,
  title={Gaussiancube: A structured and explicit radiance representation for 3d generative modeling},
  author={Zhang, Bowen and Cheng, Yiji and Yang, Jiaolong and Wang, Chunyu and Zhao, Feng and Tang, Yansong and Chen, Dong and Guo, Baining},
  journal={arXiv preprint arXiv:2403.19655},
  year={2024}
}

@inproceedings{rombach2022high,
  title={High-resolution image synthesis with latent diffusion models},
  author={Rombach, Robin and Blattmann, Andreas and Lorenz, Dominik and Esser, Patrick and Ommer, Bj{\"o}rn},
  booktitle={Proceedings of the IEEE/CVF conference on computer vision and pattern recognition},
  pages={10684--10695},
  year={2022}
}

@article{vahdat2022lion,
  title={Lion: Latent point diffusion models for 3d shape generation},
  author={Vahdat, Arash and Williams, Francis and Gojcic, Zan and Litany, Or and Fidler, Sanja and Kreis, Karsten and others},
  journal={Advances in Neural Information Processing Systems},
  volume={35},
  pages={10021--10039},
  year={2022}
}

@inproceedings{ren2024xcube,
  title={Xcube: Large-scale 3d generative modeling using sparse voxel hierarchies},
  author={Ren, Xuanchi and Huang, Jiahui and Zeng, Xiaohui and Museth, Ken and Fidler, Sanja and Williams, Francis},
  booktitle={Proceedings of the IEEE/CVF conference on computer vision and pattern recognition},
  pages={4209--4219},
  year={2024}
}

@inproceedings{chen20253dtopia,
  title={3dtopia-xl: Scaling high-quality 3d asset generation via primitive diffusion},
  author={Chen, Zhaoxi and Tang, Jiaxiang and Dong, Yuhao and Cao, Ziang and Hong, Fangzhou and Lan, Yushi and Wang, Tengfei and Xie, Haozhe and Wu, Tong and Saito, Shunsuke and others},
  booktitle={Proceedings of the Computer Vision and Pattern Recognition Conference},
  pages={26576--26586},
  year={2025}
}

@article{zhang2024clay,
  title={Clay: A controllable large-scale generative model for creating high-quality 3d assets},
  author={Zhang, Longwen and Wang, Ziyu and Zhang, Qixuan and Qiu, Qiwei and Pang, Anqi and Jiang, Haoran and Yang, Wei and Xu, Lan and Yu, Jingyi},
  journal={ACM Transactions on Graphics (TOG)},
  volume={43},
  number={4},
  pages={1--20},
  year={2024},
  publisher={ACM New York, NY, USA}
}

@article{xu2023tandem3dactivetactileexploration,
  title={TANDEM3D: Active Tactile Exploration for 3D Object Recognition},
  author={Jingxi Xu and Han Lin and Shuran Song and Matei T. Ciocarlie},
  journal={2023 IEEE International Conference on Robotics and Automation (ICRA)},
  year={2022},
  pages={10401-10407},
  url={https://api.semanticscholar.org/CorpusID:252368205}
}

@misc{zheng2024bayesianframeworkactivetactile,
      title={A Bayesian Framework for Active Tactile Object Recognition, Pose Estimation and Shape Transfer Learning}, 
      author={Haodong Zheng and Andrei Jalba and Raymond H. Cuijpers and Wijnand IJsselsteijn and Sanne Schoenmakers},
      year={2024},
      eprint={2409.06912},
      archivePrefix={arXiv},
      primaryClass={cs.RO},
      url={https://arxiv.org/abs/2409.06912}, 
}

@inproceedings{wang20183d,
  title={3d shape perception from monocular vision, touch, and shape priors},
  author={Wang, Shaoxiong and Wu, Jiajun and Sun, Xingyuan and Yuan, Wenzhen and Freeman, William T and Tenenbaum, Joshua B and Adelson, Edward H},
  booktitle={2018 IEEE/RSJ International Conference on Intelligent Robots and Systems (IROS)},
  pages={1606--1613},
  year={2018},
  organization={IEEE}
}

@article{smith20203d,
  title={3d shape reconstruction from vision and touch},
  author={Smith, Edward and Calandra, Roberto and Romero, Adriana and Gkioxari, Georgia and Meger, David and Malik, Jitendra and Drozdzal, Michal},
  journal={Advances in Neural Information Processing Systems},
  volume={33},
  pages={14193--14206},
  year={2020}
}

@inproceedings{chen2023sliding,
  title={Sliding touch-based exploration for modeling unknown object shape with multi-fingered hands},
  author={Chen, Yiting and Tekden, Ahmet Ercan and Deisenroth, Marc Peter and Bekiroglu, Yasemin},
  booktitle={2023 IEEE/RSJ International Conference on Intelligent Robots and Systems (IROS)},
  pages={8943--8950},
  year={2023},
  organization={IEEE}
}

@article{smith2021active,
  title={Active 3d shape reconstruction from vision and touch},
  author={Smith, Edward and Meger, David and Pineda, Luis and Calandra, Roberto and Malik, Jitendra and Romero Soriano, Adriana and Drozdzal, Michal},
  journal={Advances in Neural Information Processing Systems},
  volume={34},
  pages={16064--16078},
  year={2021}
}

@inproceedings{tahoun2021visual,
  title={Visual-tactile fusion for 3D objects reconstruction from a single depth view and a single gripper touch for robotics tasks},
  author={Tahoun, Mohamed and Tahri, Omar and Ram{\'o}n, Juan Antonio Corrales and Mezouar, Youcef},
  booktitle={2021 IEEE/RSJ International Conference on Intelligent Robots and Systems (IROS)},
  pages={6786--6793},
  year={2021},
  organization={IEEE}
}

@inproceedings{suresh2022shapemap,
  title={Shapemap 3-d: Efficient shape mapping through dense touch and vision},
  author={Suresh, Sudharshan and Si, Zilin and Mangelson, Joshua G and Yuan, Wenzhen and Kaess, Michael},
  booktitle={2022 International Conference on Robotics and Automation (ICRA)},
  pages={7073--7080},
  year={2022},
  organization={IEEE}
}

@inproceedings{swann2024touch,
  title={Touch-gs: Visual-tactile supervised 3d gaussian splatting},
  author={Swann, Aiden and Strong, Matthew and Do, Won Kyung and Camps, Gadiel Sznaier and Schwager, Mac and Kennedy, Monroe},
  booktitle={2024 IEEE/RSJ International Conference on Intelligent Robots and Systems (IROS)},
  pages={10511--10518},
  year={2024},
  organization={IEEE}
}

@article{luo2025tactile,
  title={Tactile robotics: An outlook},
  author={Luo, Shan and Lepora, Nathan F and Yuan, Wenzhen and Althoefer, Kaspar and Cheng, Gordon and Dahiya, Ravinder},
  journal={IEEE Transactions on Robotics},
  year={2025},
  publisher={IEEE}
}

@ARTICLE{10851808,
  author={Caddeo, Gabriele M. and Maracani, Andrea and Alfano, Paolo D. and Piga, Nicola A. and Rosasco, Lorenzo and Natale, Lorenzo},
  journal={IEEE Sensors Journal}, 
  title={Sim2Surf: A Sim2Real Surface Classifier for Vision-Based Tactile Sensors With a Bilevel Adaptation Pipeline}, 
  year={2025},
  volume={25},
  number={5},
  pages={8697-8709},
  doi={10.1109/JSEN.2025.3530712}}

@INPROCEEDINGS{11127723,
  author={Shahidzadeh, Amir-Hossein and Caddeo, Gabriele M. and Alapati, Koushik and Natale, Lorenzo and Fermüler, Cornelia and Aloimonos, Yiannis},
  booktitle={2025 IEEE International Conference on Robotics and Automation (ICRA)}, 
  title={FeelAnyForce: Estimating Contact Force Feedback from Tactile Sensation for Vision-Based Tactile Sensors}, 
  year={2025},
  volume={},
  number={},
  pages={251-257},
  doi={10.1109/ICRA55743.2025.11127723}}

@INPROCEEDINGS{8460494,
  author={Luo, Shan and Yuan, Wenzhen and Adelson, Edward and Cohn, Anthony G. and Fuentes, Raul},
  booktitle={2018 IEEE International Conference on Robotics and Automation (ICRA)}, 
  title={ViTac: Feature Sharing Between Vision and Tactile Sensing for Cloth Texture Recognition}, 
  year={2018},
  volume={},
  number={},
  pages={2722-2727},
  doi={10.1109/ICRA.2018.8460494}}

@INPROCEEDINGS{9811953,
  author={Sodhi, Paloma and Kaess, Michael and Mukadanr, Mustafa and Anderson, Stuart},
  booktitle={2022 International Conference on Robotics and Automation (ICRA)}, 
  title={PatchGraph: In-hand tactile tracking with learned surface normals}, 
  year={2022},
  volume={},
  number={},
  pages={2164-2170},
  doi={10.1109/ICRA46639.2022.9811953}}

@article{jiang2022shallitouchvisionguided,
  title={Where Shall I Touch? Vision-Guided Tactile Poking for Transparent Object Grasping},
  author={Jiaqi Jiang and Guanqun Cao and Aaron Butterworth and Thanh-Toan Do and Shan Luo},
  journal={IEEE/ASME Transactions on Mechatronics},
  year={2022},
  volume={28},
  pages={233-244},
  url={https://api.semanticscholar.org/CorpusID:251718948}
}

@inproceedings{
oller2022manipulationmembraneshighresolutionhighly,
title={Manipulation via Membranes: High-Resolution and Highly Deformable Tactile Sensing and Control},
author={Miquel Oller and Mireia Planas i Lisbona and Dmitry Berenson and Nima Fazeli},
booktitle={6th Annual Conference on Robot Learning},
year={2022},
url={https://openreview.net/forum?id=ft8IeFe4-8e}
}

@INPROCEEDINGS{wei2025mathcaldrograspunifiedrepresentation,
  author={Wei, Zhenyu and Xu, Zhixuan and Guo, Jingxiang and Hou, Yiwen and Gao, Chongkai and Cai, Zhehao and Luo, Jiayu and Shao, Lin},
  booktitle={2025 IEEE International Conference on Robotics and Automation (ICRA)}, 
  title={$\mathcal{D}(\mathcal{R}, \mathcal{O})$ Grasp: A Unified Representation of Robot and Object Interaction for Cross-Embodiment Dexterous Grasping}, 
  year={2025},
  volume={},
  number={},
  pages={4982-4988},
  doi={10.1109/ICRA55743.2025.11127754}}

@article {Wang-Sig2022,
  title      = {Dual Octree Graph Networks for Learning Adaptive Volumetric
                Shape Representations},
  author     = {Wang, Peng-Shuai and Liu, Yang and Tong, Xin},
  journal    = {ACM Transactions on Graphics (SIGGRAPH)},
  volume     = {41},
  number     = {4},
  year       = {2022},
}

@inproceedings{2016-jamali-active,
  author = {Jamali, N. and Ciliberto, C. and Rosasco, L. and Natale, L.},
  title = {Active Perception: Building Objects' Models Using Tactile Exploration},
  journal = {IEEE-RAS International Conference on Humanoid Robots},
  year = {2016},
  pages = {179-185},
  doi = {10.1109/HUMANOIDS.2016.7803275}
}

@article{2017-vezzani-memory,
  author = {Vezzani, G. and Pattacini, U. and Battistelli, G. and Chisci, L. and Natale, L.},
  title = {Memory Unscented Particle Filter for 6-DOF Tactile Localization},
  journal = {IEEE Transactions on Robotics},
  year = {2017},
  volume = {33},
  number = {5},
  pages = {1139-1155},
  doi = {10.1109/TRO.2017.2707092}
}

@ARTICLE{7254318,
  author={Calli, Berk and Walsman, Aaron and Singh, Arjun and Srinivasa, Siddhartha and Abbeel, Pieter and Dollar, Aaron M.},
  journal={IEEE Robotics Automation Magazine}, 
  title={Benchmarking in Manipulation Research: Using the Yale-CMU-Berkeley Object and Model Set}, 
  year={2015},
  volume={22},
  number={3},
  pages={36-52},
  doi={10.1109/MRA.2015.2448951}}

@article{wang2024training,
  title={Training free guided flow matching with optimal control},
  author={Wang, Luran and Cheng, Chaoran and Liao, Yizhen and Qu, Yanru and Liu, Ge},
  journal={arXiv preprint arXiv:2410.18070},
  year={2024}
}

@inproceedings{li2025dso,
  title={Dso: Aligning 3d generators with simulation feedback for physical soundness},
  author={Li, Ruining and Zheng, Chuanxia and Rupprecht, Christian and Vedaldi, Andrea},
  booktitle={Proceedings of the IEEE/CVF International Conference on Computer Vision},
  pages={6772--6783},
  year={2025}
}

@article{fu20213d,
  title={3d-future: 3d furniture shape with texture},
  author={Fu, Huan and Jia, Rongfei and Gao, Lin and Gong, Mingming and Zhao, Binqiang and Maybank, Steve and Tao, Dacheng},
  journal={International Journal of Computer Vision},
  volume={129},
  number={12},
  pages={3313--3337},
  year={2021},
  publisher={Springer}
}

@inproceedings{khanna2024habitat,
  title={Habitat synthetic scenes dataset (hssd-200): An analysis of 3d scene scale and realism tradeoffs for objectgoal navigation},
  author={Khanna, Mukul and Mao, Yongsen and Jiang, Hanxiao and Haresh, Sanjay and Shacklett, Brennan and Batra, Dhruv and Clegg, Alexander and Undersander, Eric and Chang, Angel X and Savva, Manolis},
  booktitle={Proceedings of the IEEE/CVF Conference on Computer Vision and Pattern Recognition},
  pages={16384--16393},
  year={2024}
}

@inproceedings{collins2022abo,
  title={Abo: Dataset and benchmarks for real-world 3d object understanding},
  author={Collins, Jasmine and Goel, Shubham and Deng, Kenan and Luthra, Achleshwar and Xu, Leon and Gundogdu, Erhan and Zhang, Xi and Vicente, Tomas F Yago and Dideriksen, Thomas and Arora, Himanshu and others},
  booktitle={Proceedings of the IEEE/CVF conference on computer vision and pattern recognition},
  pages={21126--21136},
  year={2022}
}

@inproceedings{downs2022google,
  title={Google scanned objects: A high-quality dataset of 3d scanned household items},
  author={Downs, Laura and Francis, Anthony and Koenig, Nate and Kinman, Brandon and Hickman, Ryan and Reymann, Krista and McHugh, Thomas B and Vanhoucke, Vincent},
  booktitle={2022 International Conference on Robotics and Automation (ICRA)},
  pages={2553--2560},
  year={2022},
  organization={Ieee}
}

@article{deitke2023objaverse,
  title={Objaverse-xl: A universe of 10m+ 3d objects},
  author={Deitke, Matt and Liu, Ruoshi and Wallingford, Matthew and Ngo, Huong and Michel, Oscar and Kusupati, Aditya and Fan, Alan and Laforte, Christian and Voleti, Vikram and Gadre, Samir Yitzhak and others},
  journal={Advances in Neural Information Processing Systems},
  volume={36},
  pages={35799--35813},
  year={2023}
}

@inproceedings{lin2018learning,
  title={Learning efficient point cloud generation for dense 3d object reconstruction},
  author={Lin, Chen-Hsuan and Kong, Chen and Lucey, Simon},
  booktitle={proceedings of the AAAI Conference on Artificial Intelligence},
  volume={32},
  number={1},
  year={2018}
}

@inproceedings{chan2022efficient,
  title={Efficient geometry-aware 3d generative adversarial networks},
  author={Chan, Eric R and Lin, Connor Z and Chan, Matthew A and Nagano, Koki and Pan, Boxiao and De Mello, Shalini and Gallo, Orazio and Guibas, Leonidas J and Tremblay, Jonathan and Khamis, Sameh and others},
  booktitle={Proceedings of the IEEE/CVF conference on computer vision and pattern recognition},
  pages={16123--16133},
  year={2022}
}

@inproceedings{niemeyer2021giraffe,
  title={Giraffe: Representing scenes as compositional generative neural feature fields},
  author={Niemeyer, Michael and Geiger, Andreas},
  booktitle={Proceedings of the IEEE/CVF conference on computer vision and pattern recognition},
  pages={11453--11464},
  year={2021}
}

@article{schwarz2020graf,
  title={Graf: Generative radiance fields for 3d-aware image synthesis},
  author={Schwarz, Katja and Liao, Yiyi and Niemeyer, Michael and Geiger, Andreas},
  journal={Advances in neural information processing systems},
  volume={33},
  pages={20154--20166},
  year={2020}
}

@inproceedings{melas2023pc2,
  title={Pc2: Projection-conditioned point cloud diffusion for single-image 3d reconstruction},
  author={Melas-Kyriazi, Luke and Rupprecht, Christian and Vedaldi, Andrea},
  booktitle={Proceedings of the IEEE/CVF conference on computer vision and pattern recognition},
  pages={12923--12932},
  year={2023}
}

@inproceedings{wu2023sketch,
  title={Sketch and text guided diffusion model for colored point cloud generation},
  author={Wu, Zijie and Wang, Yaonan and Feng, Mingtao and Xie, He and Mian, Ajmal},
  booktitle={Proceedings of the IEEE/CVF International Conference on Computer Vision},
  pages={8929--8939},
  year={2023}
}

@inproceedings{li2023diffusion,
  title={Diffusion-sdf: Text-to-shape via voxelized diffusion},
  author={Li, Muheng and Duan, Yueqi and Zhou, Jie and Lu, Jiwen},
  booktitle={Proceedings of the IEEE/CVF conference on computer vision and pattern recognition},
  pages={12642--12651},
  year={2023}
}

@inproceedings{chen2023single,
  title={Single-stage diffusion nerf: A unified approach to 3d generation and reconstruction},
  author={Chen, Hansheng and Gu, Jiatao and Chen, Anpei and Tian, Wei and Tu, Zhuowen and Liu, Lingjie and Su, Hao},
  booktitle={Proceedings of the IEEE/CVF international conference on computer vision},
  pages={2416--2425},
  year={2023}
}

@inproceedings{shue20233d,
  title={3d neural field generation using triplane diffusion},
  author={Shue, J Ryan and Chan, Eric Ryan and Po, Ryan and Ankner, Zachary and Wu, Jiajun and Wetzstein, Gordon},
  booktitle={Proceedings of the IEEE/CVF conference on computer vision and pattern recognition},
  pages={20875--20886},
  year={2023}
}

@inproceedings{zhang2024rodinhd,
  title={Rodinhd: High-fidelity 3d avatar generation with diffusion models},
  author={Zhang, Bowen and Cheng, Yiji and Wang, Chunyu and Zhang, Ting and Yang, Jiaolong and Tang, Yansong and Zhao, Feng and Chen, Dong and Guo, Baining},
  booktitle={European Conference on Computer Vision},
  pages={465--483},
  year={2024},
  organization={Springer}
}

@inproceedings{
albergo2022building,
title={Building Normalizing Flows with Stochastic Interpolants},
author={Michael Samuel Albergo and Eric Vanden-Eijnden},
url={https://arxiv.org/abs/2209.15571},
booktitle={The Eleventh International Conference on Learning Representations },
year={2023},
}

@inproceedings{
lipman2022flow,
title={Flow Matching for Generative Modeling},
author={Yaron Lipman and Ricky T. Q. Chen and Heli Ben-Hamu and Maximilian Nickel and Matthew Le},
booktitle={The Eleventh International Conference on Learning Representations },
year={2023},
url={https://openreview.net/forum?id=PqvMRDCJT9t}
}

@article{liu2022flow,
  title={Flow straight and fast: Learning to generate and transfer data with rectified flow},
  author={Liu, Xingchao and Gong, Chengyue and Liu, Qiang},
  journal={arXiv preprint arXiv:2209.03003},
  year={2022}
}

@misc{tang2024automatespecialistgeneralistassembly,
      title={AutoMate: Specialist and Generalist Assembly Policies over Diverse Geometries}, 
      author={Bingjie Tang and Iretiayo Akinola and Jie Xu and Bowen Wen and Ankur Handa and Karl Van Wyk and Dieter Fox and Gaurav S. Sukhatme and Fabio Ramos and Yashraj Narang},
      year={2024},
      eprint={2407.08028},
      archivePrefix={arXiv},
      primaryClass={cs.RO},
      url={https://arxiv.org/abs/2407.08028}, 
}

@misc{wen2024foundationposeunified6dpose,
      title={FoundationPose: Unified 6D Pose Estimation and Tracking of Novel Objects}, 
      author={Bowen Wen and Wei Yang and Jan Kautz and Stan Birchfield},
      year={2024},
      eprint={2312.08344},
      archivePrefix={arXiv},
      primaryClass={cs.CV},
      url={https://arxiv.org/abs/2312.08344}, 
}

@misc{oquab2024dinov2learningrobustvisual,
      title={DINOv2: Learning Robust Visual Features without Supervision}, 
      author={Maxime Oquab and Timothée Darcet and Théo Moutakanni and Huy Vo and Marc Szafraniec and Vasil Khalidov and Pierre Fernandez and Daniel Haziza and Francisco Massa and Alaaeldin El-Nouby and Mahmoud Assran and Nicolas Ballas and Wojciech Galuba and Russell Howes and Po-Yao Huang and Shang-Wen Li and Ishan Misra and Michael Rabbat and Vasu Sharma and Gabriel Synnaeve and Hu Xu and Hervé Jegou and Julien Mairal and Patrick Labatut and Armand Joulin and Piotr Bojanowski},
      year={2024},
      eprint={2304.07193},
      archivePrefix={arXiv},
      primaryClass={cs.CV},
      url={https://arxiv.org/abs/2304.07193}, 
}

@misc{xiang2018posecnnconvolutionalneuralnetwork,
      title={PoseCNN: A Convolutional Neural Network for 6D Object Pose Estimation in Cluttered Scenes}, 
      author={Yu Xiang and Tanner Schmidt and Venkatraman Narayanan and Dieter Fox},
      year={2018},
      eprint={1711.00199},
      archivePrefix={arXiv},
      primaryClass={cs.CV},
      url={https://arxiv.org/abs/1711.00199}, 
}

@inproceedings{yang2024physcene,
  title={Physcene: Physically interactable 3d scene synthesis for embodied ai},
  author={Yang, Yandan and Jia, Baoxiong and Zhi, Peiyuan and Huang, Siyuan},
  booktitle={Proceedings of the IEEE/CVF Conference on Computer Vision and Pattern Recognition},
  pages={16262--16272},
  year={2024}
}

@article{ni2024phyrecon,
  title={Phyrecon: Physically plausible neural scene reconstruction},
  author={Ni, Junfeng and Chen, Yixin and Jing, Bohan and Jiang, Nan and Wang, Bin and Dai, Bo and Li, Puhao and Zhu, Yixin and Zhu, Song-Chun and Huang, Siyuan},
  journal={Advances in Neural Information Processing Systems},
  volume={37},
  pages={25747--25780},
  year={2024}
}

@article{chen2024atlas3d,
  title={Atlas3d: Physically constrained self-supporting text-to-3d for simulation and fabrication},
  author={Chen, Yunuo and Xie, Tianyi and Zong, Zeshun and Li, Xuan and Gao, Feng and Yang, Yin and Wu, Ying Nian and Jiang, Chenfanfu},
  journal={Advances in Neural Information Processing Systems},
  volume={37},
  pages={102501--102530},
  year={2024}
}

@article{guo2024physically,
  title={Physically compatible 3d object modeling from a single image},
  author={Guo, Minghao and Wang, Bohan and Ma, Pingchuan and Zhang, Tianyuan and Owens, Crystal and Gan, Chuang and Tenenbaum, Josh and He, Kaiming and Matusik, Wojciech},
  journal={Advances in Neural Information Processing Systems},
  volume={37},
  pages={119260--119282},
  year={2024}
}

@inproceedings{ye2022hand,
    author = {Ye, Yufei
              and Gupta, Abhinav
              and Tulsiani, Shubham},
    title = {What's in your hands? 3D Reconstruction of Generic Objects in Hands},
    booktitle = {CVPR},
    year = {2022}
}

@inproceedings{petrov2023popup,
   title={Object pop-up: Can we infer 3D objects and their poses from human interactions alone?},
   author={Petrov, Ilya A and Marin, Riccardo and Chibane, Julian and Pons-Moll, Gerard},
   booktitle={Proceedings of the IEEE/CVF Conference on Computer Vision and Pattern Recognition},
   year={2023}
}

@InProceedings{chen2022alignsdf,
author       = {Chen, Zerui and Hasson, Yana and Schmid, Cordelia and Laptev, Ivan},
title        = {{AlignSDF}: {Pose-Aligned} Signed Distance Fields for Hand-Object Reconstruction},
booktitle    = {ECCV},
year         = {2022},
}

@INPROCEEDINGS {easyHOI,
author = { Liu, Yumeng and Long, Xiaoxiao and Yang, Zemin and Liu, Yuan and Habermann, Marc and Theobalt, Christian and Ma, Yuexin and Wang, Wenping },
booktitle = { 2025 IEEE/CVF Conference on Computer Vision and Pattern Recognition (CVPR) },
title = {{ EasyHOI: Unleashing the Power of Large Models for Reconstructing Hand-Object Interactions in the Wild }},
year = {2025},
volume = {},
ISSN = {},
pages = {7037-7047},
doi = {10.1109/CVPR52734.2025.00660},
url = {https://doi.ieeecomputersociety.org/10.1109/CVPR52734.2025.00660},
publisher = {IEEE Computer Society},
address = {Los Alamitos, CA, USA},
month =Jun}

@article{aytekin2025follow,
  title={Follow My Hold: Hand-object Interaction Reconstruction through Geometric Guidance},
  author={Aytekin, Ayce Idil and Rhodin, Helge and Dabral, Rishabh and Theobalt, Christian},
  year={2026},
  journal={International Conference on 3D Vision},
}

@article{manipulatorIJRR,
author = {Krainin, Michael and Henry, Peter and Ren, Xiaofeng and Fox, Dieter},
year = {2011},
month = {10},
pages = {1311-1327},
title = {Manipulator and object tracking for in-hand 3D object modeling},
volume = {30},
journal = {I. J. Robotic Res.},
doi = {10.1177/0278364911403178}
}

@misc{zheng2026bayesianactiveobjectrecognition,
      title={Bayesian Active Object Recognition and 6D Pose Estimation from Multimodal Contact Sensing}, 
      author={Haodong Zheng and Gabriele M. Caddeo and Andrei C. Jalba and Wijnand A. IJsselsteijn and Lorenzo Natale and Raymond H. Cuijpers},
      year={2026},
      eprint={2603.21410},
      archivePrefix={arXiv},
      primaryClass={cs.RO},
      url={https://arxiv.org/abs/2603.21410}, 
}

@INPROCEEDINGS{9578786,
  author={Chao, Y.-W., et al.},
  booktitle={CVPR 2021}, 
  title={DexYCB: A Benchmark for Capturing Hand Grasping of Objects}, 
  year={2021},
  volume={},
  number={},
  pages={9040-9049},
}

@INPROCEEDINGS{10160359,
  author={Caddeo, Gabriele M. and Piga, Nicola A. and Bottarel, Fabrizio and Natale, Lorenzo},
  booktitle={2023 IEEE International Conference on Robotics and Automation (ICRA)}, 
  title={Collision-aware In-hand 6D Object Pose Estimation using Multiple Vision-based Tactile Sensors}, 
  year={2023},
  volume={},
  number={},
  pages={719-725},
  doi={10.1109/ICRA48891.2023.10160359}}
\newpage
\title{  Supplementary Materials for Physically Grounded 3D Generative Reconstruction under Hand Occlusion using Proprioception and Multi-Contact Touch
}
\titlerunning{Physically Grounded 3D Generative
Reconstruction}

\author{Gabriele M. Caddeo\inst{1}\orcidlink{0000-0003-1566-1314} \and
Pasquale Marra\inst{1}\orcidlink{0009-0002-7110-1952} \and
Lorenzo Natale\inst{1}\orcidlink{0000-0002-8777-5233}}

\authorrunning{G. M. ~Caddeo et al.}

\institute{
Istituto Italiano di Tecnologia\\
\email{\{gabriele.caddeo,pasquale.marra,lorenzo.natale\}@iit.it}\\
\url{https://hsp.iit.it/}
}
\maketitle

\appendix
\setcounter{table}{0}
\renewcommand{\thetable}{A\arabic{table}}
\setcounter{figure}{0}
\renewcommand{\thefigure}{A\arabic{figure}}

\section{More details on Datasets} \label{appendix:dataset}
\subsection{SDF Generation}
The SDFs are generated by first normalizing the mesh geometry into a shared canonical domain, $\Omega=[-1,1]^3$, using a common scale and center for the hand--object scene. A canonical signed distance field is then computed on a regular grid using the open-source code from~\cite{Wang-Sig2022}. For each frame, the hand and object meshes are transformed with the same pose-dependent similarity transform, including rotation, scale, and translation, so that the full scene remains consistently aligned in the voxel grid. The canonical SDF is resampled into the posed grid by inverse mapping grid centers through the applied transform, while the sign is recovered from the posed mesh using winding numbers. Near the surface, distances are refined with exact point-to-mesh queries to improve accuracy in the narrow band around the zero level set. The final SDFs are stored on a fixed-resolution grid, with negative values inside the mesh and positive values outside.
\subsection{Image Processing}
\paragraph{Pose-consistent sprite placement.}
For each instance and camera view, we start from a rendered RGBA image $I^{\mathrm{rgba}}$ and pose metadata extracted from the posed 3D grid. The metadata provides the grid resolution $R$ and a 2D bounding box $\mathrm{bbox}_{xy}=(x_{\min},y_{\min},x_{\max},y_{\max})$ defined in \emph{grid-index} coordinates on the $(x,y)$ plane, corresponding to the footprint/projection of the posed object in the canonical grid.
We embed the rendered object into a square conditioning canvas of size $H\times W$ (with $H=W$) so that its 2D translation and scale match the object placement in the 3D grid.
We convert the voxel-space bounding box to pixel coordinates using
\begin{equation}
s_{\mathrm{px}}=\frac{W}{R},
\end{equation}
which yields a target pixel box in the canvas.
We then extract a tight sprite crop from $I^{\mathrm{rgba}}$ using its alpha channel (the smallest rectangle covering all nonzero-alpha pixels), resize the sprite with a uniform scale factor that fits it inside the target pixel box while preserving aspect ratio, and paste it centered within the box.
Resizing uses a high-quality filter (e.g., Lanczos) to avoid aliasing.

\begin{figure}[t]
  \centering

  \includegraphics[width=\textwidth]{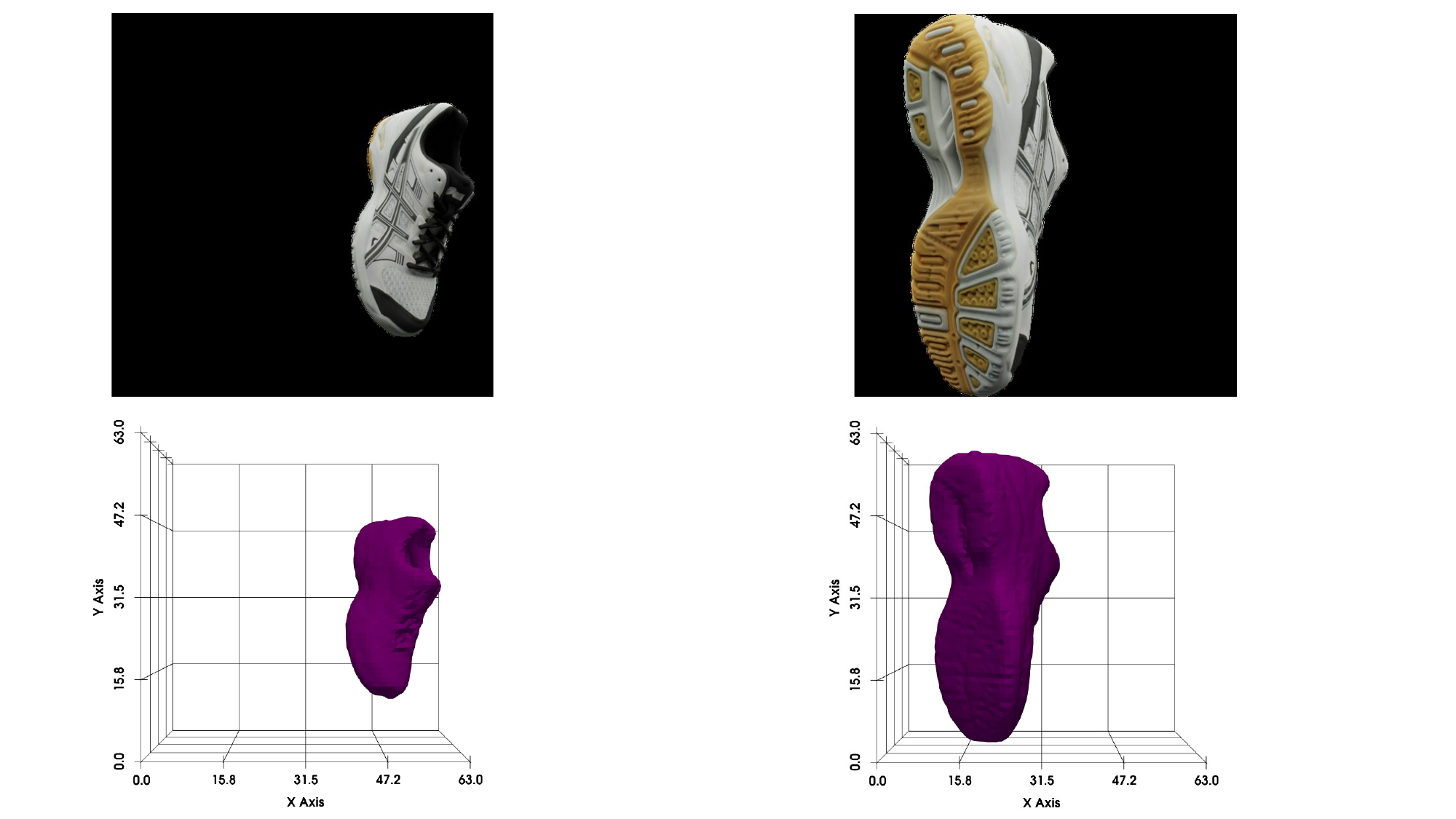}
  \caption{\textbf{Image processing}. The conditioning images are processed to be pose consistent with the 0-level of the SDF.}
  \label{fig:image_processing}
\end{figure}

\paragraph{Conditioning image construction.}
After compositing the resized sprite onto the canvas, we obtain an RGBA image $\widetilde{I}^{\mathrm{rgba}}$ and convert it to an RGB conditioning tensor by premultiplying alpha over a black background:
\begin{equation}
I = \widetilde{I}^{\mathrm{rgb}} \odot \widetilde{I}^{\alpha},
\end{equation}
where $\widetilde{I}^{\mathrm{rgb}}\in[0,1]^{H\times W\times 3}$ and $\widetilde{I}^{\alpha}\in[0,1]^{H\times W\times 1}$ are the RGB and alpha channels of $\widetilde{I}^{\mathrm{rgba}}$ and $\odot$ denotes elementwise multiplication (broadcast over channels). This yields a pose-consistent image $I\in[0,1]^{3\times H\times W}$. ~\cref{fig:image_processing} shows an example of image construction.
\paragraph{Aligned occlusion masks (when present).}
In the occluded setting, we additionally have a hand/occluder mask and an object mask for the same view, denoted by $M_h$ and $M_o$.
To guarantee pixel-accurate alignment with the conditioning image, we apply the \emph{same} crop, resize, and paste parameters used for the RGBA sprite to both masks: each mask is cropped with the same sprite crop window, resized to the same target size (using nearest-neighbor interpolation to preserve sharp boundaries), and pasted at the same location on a blank canvas.
This produces aligned masks $M_h,M_o\in[0,1]^{H\times W}$ and the visible-object conditioning image
\begin{equation}
I_o = I \odot M_o,
\end{equation}
where $M_o$ is broadcast over RGB channels.
The model therefore receives pose-consistent visual evidence ($I$ or $I_o$) together with explicit occlusion cues $(M_o,M_h)$.

\subsection{Grasp Simulation Dataset}
\label{appendix:dataset_sim}

To generate grasp scenes for finetuning, we use a pretrained grasp-synthesis model from~\cite{wei2025mathcaldrograspunifiedrepresentation}, which supports three end-effectors: Barrett, ShadowHand, and Allegro. These robotic manipulators feature $3$, $4$, and $5$ fingers respectively, offering a good generalization in terms of morphology and shape.
Because grasp quality depends on object scale, we restrict training objects to sizes comparable to those used to train the grasp model.
We therefore use Google Scanned Objects (GSO)~\cite{downs2022google} for training and evaluate on a fixed subset of YCB~\cite{7254318}.

\paragraph{YCB evaluation set.}
We use the following YCB objects:
\texttt{002\_master\_chef\_can},
\texttt{004\_sugar\_box},
\texttt{006\_mustard\_bottle},
\texttt{007\_tuna\_fish\_can},
\texttt{008\_pudding\_box},
\texttt{009\_gelatin\_box},
\texttt{010\_potted\_meat\_can},
\texttt{011\_banana},
\texttt{012\_strawberry},\\
\texttt{013\_apple},
\texttt{014\_lemon},
\texttt{015\_peach},
\texttt{016\_pear},
\texttt{017\_orange},
\texttt{018\_plum},\\
\texttt{019\_pitcher\_base},
\texttt{021\_bleach\_cleanser},
\texttt{022\_windex\_bottle},
\texttt{024\_bowl},\\
\texttt{025\_mug},
\texttt{026\_sponge},
\texttt{030\_fork},
\texttt{031\_spoon},
\texttt{032\_knife},
\texttt{035\_power\_drill},
\texttt{036\_wood\_block},
\texttt{037\_scissors},
\texttt{038\_padlock},
\texttt{042\_adjustable\_wrench},\\
\texttt{043\_phillips\_screwdriver},
\texttt{044\_flat\_screwdriver},
\texttt{048\_hammer},\\
\texttt{050\_medium\_clamp},
\texttt{051\_large\_clamp},
\texttt{052\_extra\_large\_clamp}.\\
For each object–hand pair, we run the grasp generator and retain up to 10 grasps to maintain a balanced and diverse dataset. A grasp is considered valid if it yields at least one contact and exhibits no hand–object interpenetration. For each valid grasp, we render $24$ images from randomized camera viewpoints around the scene. For the GSO dataset we obtain $19,413$ grasp tuples, with a total of $465,912$ images, while for the YCB dataset we obtain $945$ different grasps, with $22,680$ images rendered.

\paragraph{Scene canonicalization in the grid.}
For each simulated grasp, we voxelize and maximize both the posed hand and object into the same canonical grid, keeping a padding of $p$ voxels from the grid boundary (we use $p=2$).
We center the \emph{hand--object scene} in the grid $(x,y)$ plane, while preserving the relative object placement within the scene; consequently, the object is not necessarily centered, consistent with the pose variability used during pretraining.
Along the viewing direction ($z$), rather than centering the entire scene, we center the centroid of the fingertip set, which allocates more free space in front of the hand and better accommodates objects that extend along the camera axis while remaining within the grid domain.
 
\subsection{Real-world Dataset}
\label{appendix:dataset_real_world}
We collect real-world data on a humanoid platform equipped with a head-mounted Intel RealSense camera and XELA magnetic tactile sensors on the fingertips. The real robot hand differs from the end-effectors used during training and has five fingertips (as in ShadowHand). Using forward kinematics and a calibrated hand--camera transform, we align the hand reference frame to the camera frame such that the hand $z$ axis matches the viewing direction. For each time step, we generate the posed hand mesh consistent with joint encoder readings and convert it to an SDF grid using the same procedure adopted in simulation.

In the real setting we do not assume access to the object mesh. We therefore center the scene implicitly through hand placement: the hand is positioned so that its projected mask is consistent, in scale and translation, with its relative extent in the union mask of hand and object in the image plane. Along the viewing direction ($z$), we center using the centroid of the fingertips, as in the simulated setup, to allocate sufficient grid space for objects extending along the camera axis.

To estimate contact points from tactile measurements, we threshold taxel activations to detect active contacts. Detected contact locations are projected onto the fingertip surface and transformed into the grid frame using the same hand pose used to voxelize the hand, yielding contact coordinates consistent with the hand SDF in the canonical grid.

\section{More Details on Experiments}
\label{appendix:exp}

\subsection{Real-World Experiments}
\label{appendix:real_world}
\begin{figure}[t]
  \centering

  \includegraphics[width=\textwidth]{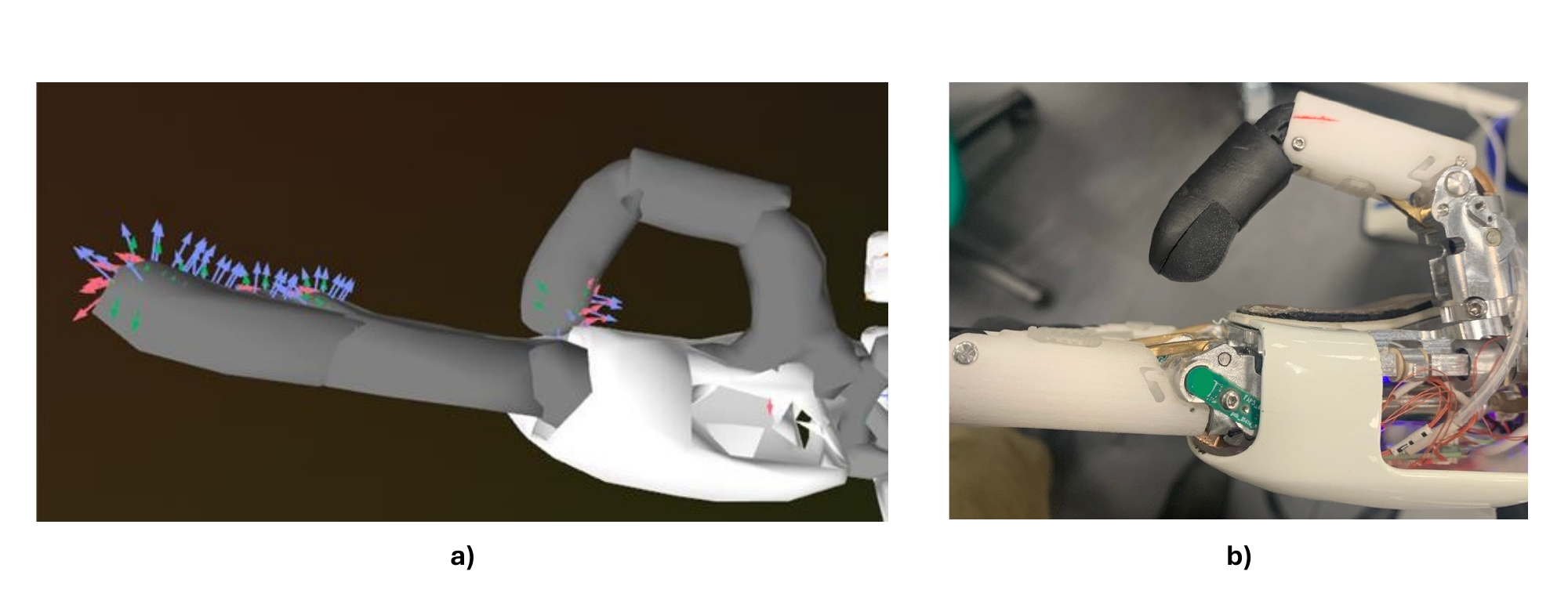}
  \caption{\textbf{Calibration error}. The figure shows a comparison between the 3D model (\textbf{a)}) and the real counterpart (\textbf{b)}) of the humanoid hand used in the real-world with a noisy calibration.  }
  \label{fig:calibration}
\end{figure}
\begin{figure}[h]
  \centering

  \includegraphics[width=\textwidth]{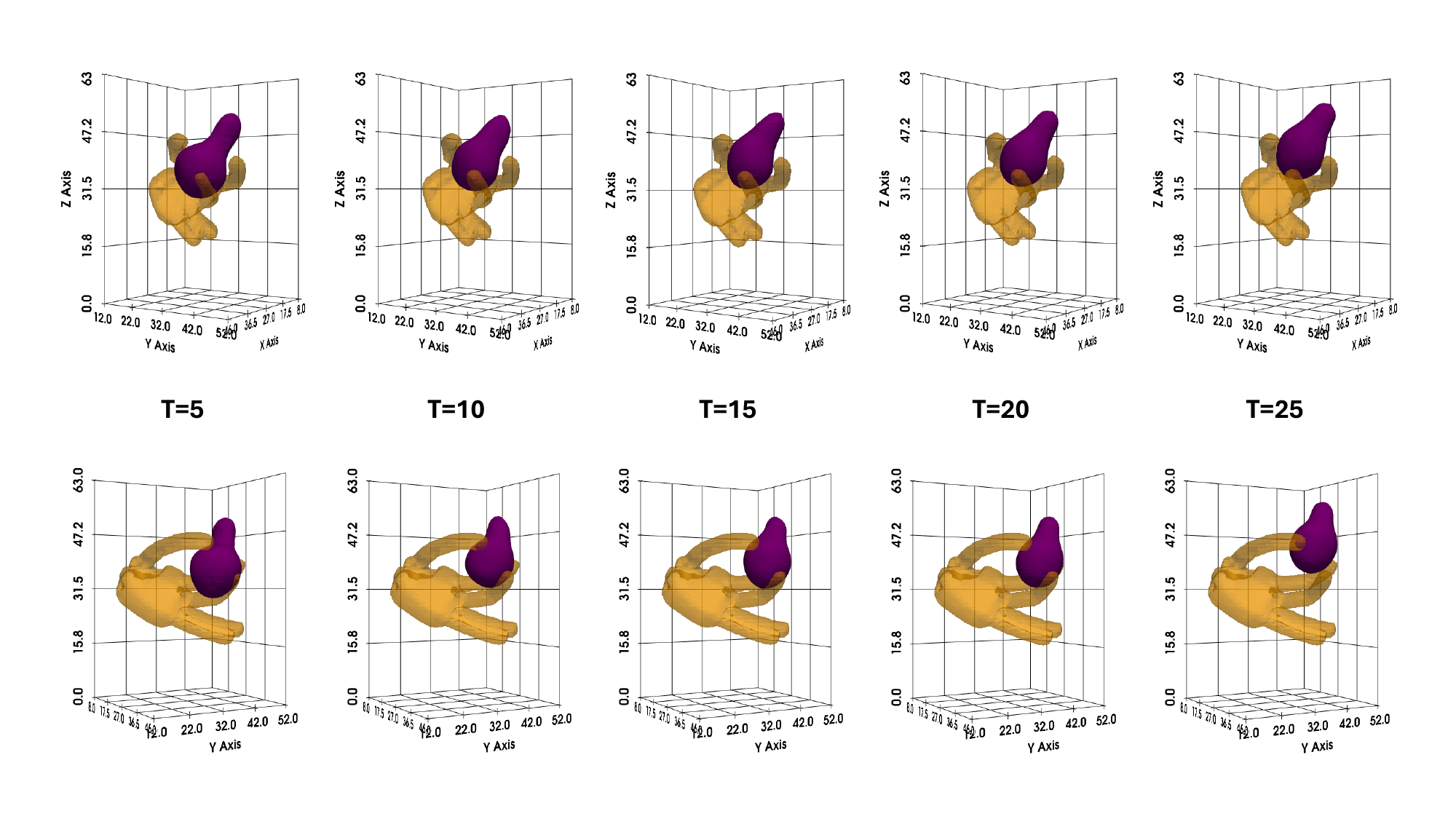}
  \caption{\textbf{Physical guidance}. During the early sampling iterations ($k<5$), decoder-based physics guidance has limited effect: the predicted SDF is still noisy, the fingertip contacts are far from the zero level set, and the object may interpenetrate the middle finger. As sampling progresses, guidance becomes effective and steers the reconstruction toward a physically compatible configuration, reducing penetration and aligning the surface with the observed contacts. The two rows show two different viewpoints of the same sampling trajectory.}
  \label{fig:guidance}
\end{figure}

We evaluate real-world transfer on five YCB objects:
\texttt{004\_sugar\_box}, \\ \texttt{006\_mustard\_bottle}, \texttt{012\_strawberry}, \texttt{016\_pear}, and  \texttt{021\_bleach\_cleanser}.
Using real counterparts of objects tested in simulation allows us to analyze both the Sim2Real gap and transfer across embodiments, as the real robot hand differs from the end-effectors used during training.
Because none of the training end-effectors are available on our platform, we do not isolate end-effector identity as a controlled variable.

A key additional source of error in the real setup is proprioception and calibration noise.
While the noise ablation in ~\cref{subsec:ablation} studies robustness to tactile perturbations, real-world deployment also depends critically on accurate camera--hand extrinsics and forward kinematics.
Imperfect calibration can lead to substantial kinematic misalignment (Fig.~\ref{fig:calibration}); even after careful calibration, small residual errors remain and effectively test robustness to noisy encoders and pose estimation. 
In practice, the model remains robust to small misalignments, and the physics guidance can partially correct them during sampling.
Figure~\ref{fig:guidance} illustrates a representative trajectory: guidance has limited effect during the early iterations, when the reconstruction is still highly noisy, but becomes effective as sampling progresses, reducing penetration (\eg, into the middle finger) and pulling the surface toward the observed contacts (\eg, at the index and thumb), resulting in a more physically compatible shape and pose.

\subsection{Comparison with Human-hand Baselines}
We further compare our method with~\cite{aytekin2025follow,easyHOI}. Unfortunately, run these methods on our data would have been intractable, as they require several minutes of inference to obtain a mesh; e.g.,~\cite{aytekin2025follow} requires more than 10 minutes per sample on an RTX A6000, whereas our method runs in tens of seconds. Moreover, their method is made specifically for human-hands, and adapting our dataset to their algorithm would require an ad-hoc retargeting. To provide a comparison, we evaluated our method on the DexYCB~\cite{9578786} test split used by~\cite{aytekin2025follow}. Table~\ref{tab:baseline} reports the results using the four metrics adopted in~\cite{aytekin2025follow}. We note that these differ from the metrics used in our paper; refer to ~\cite{aytekin2025follow} for the description. Compared with our simulated and real-world settings, the cropped images have lower resolution and the provided masks are realistically noisy. Nevertheless, our method achieves comparable (or better) results than~\cite{aytekin2025follow} and ~\cite{easyHOI}. We remark that the Reconstruction Rate (R.R.) is the 99\% against 58\% and 30\% of the baselines, and therefore our metrics are computed including those challenging samples that are discarded by~\cite{aytekin2025follow} and~\cite{easyHOI}, which averaged the results only for the reconstructed ones. Moreover, the inference time is not comparable (up to more than to 30x), and the human-hand was never seen during our training. Differently from our real-world experiments, the DexYCB dataset provide very precise poses, showing the benefits of our approach when fed with precise kinematics.

\begin{table}[ht]
    \centering
    \caption{3D reconstruction results for DexYCB~\cite{9578786} subset.}
    \label{tab:baseline}

    \resizebox{\columnwidth}{!}{%
    \begin{tabular}{c|c|c|c|c|c}
        \hline
        \textbf{Method} & \textbf{CD (cm$^2$)} $\downarrow$ & \textbf{F5mm} $\uparrow$ & \textbf{F10mm} $\uparrow$ & \textbf{I.V. (cm$^3$)} $\downarrow$ & \textbf{R.R.} $\uparrow$\\
        \hline
        \textbf{EasyHOI~\cite{easyHOI}} 
        & $ 6.26$ & $0.090$ &  $0.176$ & $19.13$ & $ 0.30$ \\
        \hline
        \textbf{FollowMyHold~\cite{aytekin2025follow}}
        & \bm{$2.04$} & $0.158$ & $0.300$ & $7.02$ & $0.58$ \\
        \hline
        \textbf{Ours}
        & $2.51$ & \bm{$0.532$} & \bm{$0.774$} & \bm{$3.27$} & \bm{$0.99$} \\
        \hline
    \end{tabular}%
    }
\end{table}

\subsection{Additional Information}
\subsubsection{Inference frequency} On an NVIDIA RTX A6000 and an H200, our method requires 11s/5s for Stage A and 7s/3s for Stage B, respectively. TRELLIS requires 5s/2s for Stage A and 5s/2s for Stage B on the same GPUs.
\subsubsection{Integration into SAM3D} Our method can be integrated into \cite{sam3dteam2025sam3d3dfyimages} by replacing the Geometry Model output with our coarse structure, rather than operating in the latent space. Our observation is supported by empirical evidence that~\cite{sam3dteam2025sam3d3dfyimages} is invariant to pose and scale; the main requirement is that the voxelization follows the expected distribution. Preliminary experiments show that this yields coherent results.
\subsubsection{Seed used for qualitative results} 
\begin{figure}[h]
  \centering
  \includegraphics[width=\textwidth]{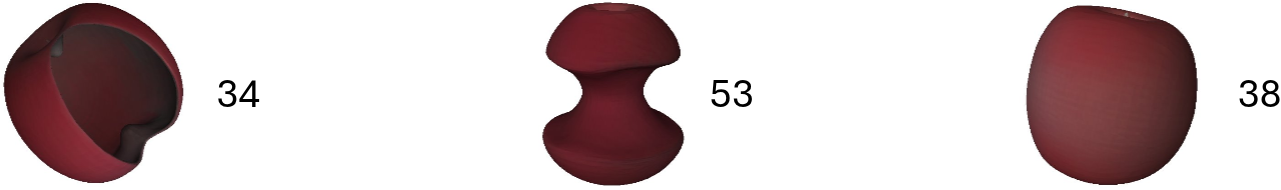}
  \caption{\textbf{Impact of seeds} Output for different seeds of Apple case in Fig. 4.} \label{fig:seeds}
\end{figure}

We acknowledge that all tested methods depend on the chosen seed, as the sampling from the source distribution affects the outcomes; for the qualitative results, we used the default seed, $42$, while for the quantitative results we average for different seeds. Following the rebuttals, we tested multiple seeds on the image in the third row of Figure~\ref{fig:real_qualitative} to show that while it's tryue that different seed can yield better results, it's still true that the output can still be poor. Indeed, although some seeds improve the reconstruction, others still fail. Fig.~\ref{fig:seeds} shows two poor reconstructions (seeds 34, 53), and a better one (seed 38). 

\subsection{Qualitative images}
\begin{figure}[h]
  \centering
  \includegraphics[width=\textwidth]{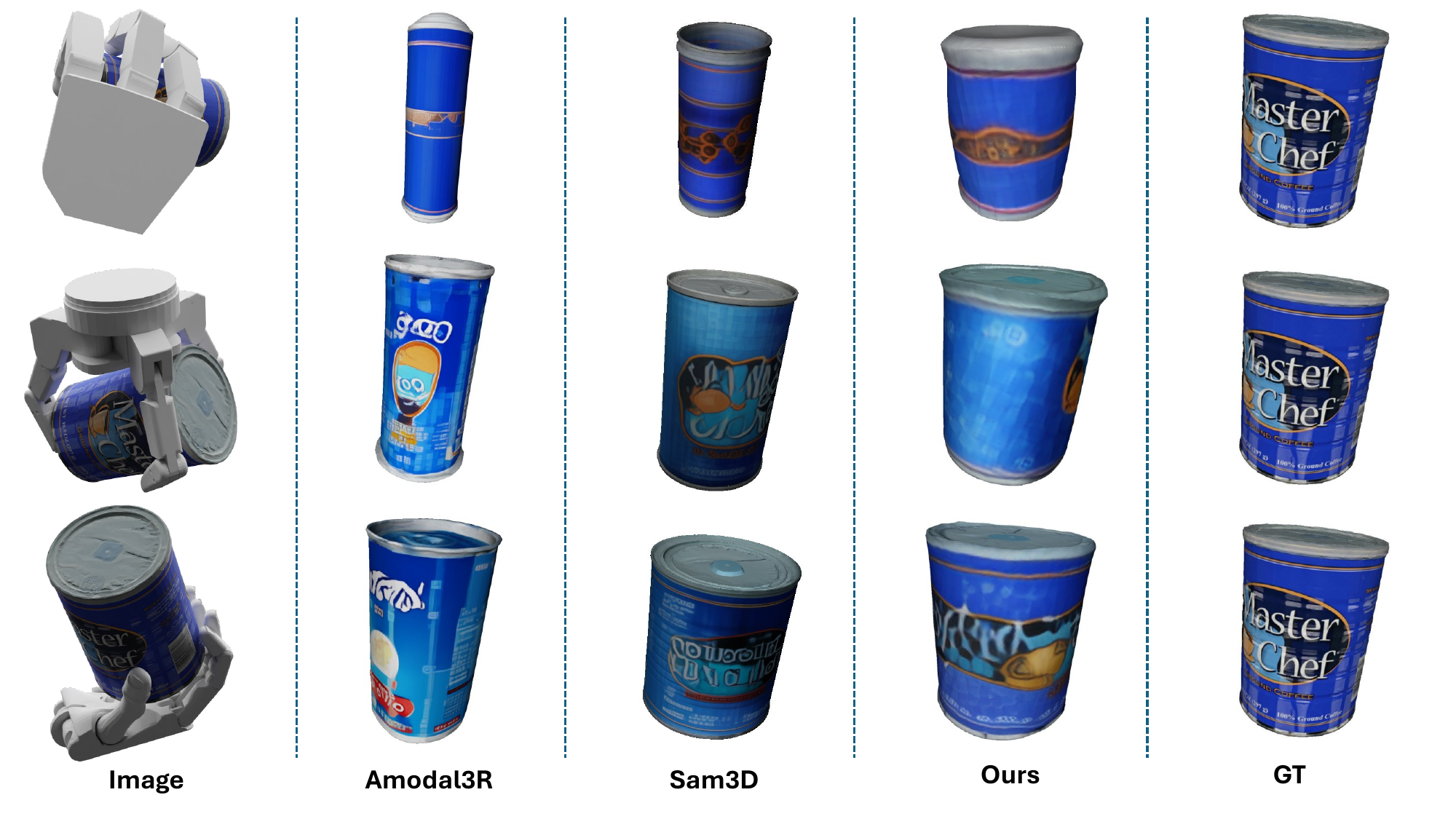}
  \caption{\textbf{Different end-effector}. Reconstructions remain consistent with the grasp geometry when changing the end-effector. As occlusion increases (bottom to top), reconstruction quality degrades, but our method preserves physically plausible relative dimensions.}
  \label{fig:qual_2}
\end{figure}

\begin{figure}[h]
  \centering
  \includegraphics[width=\textwidth]{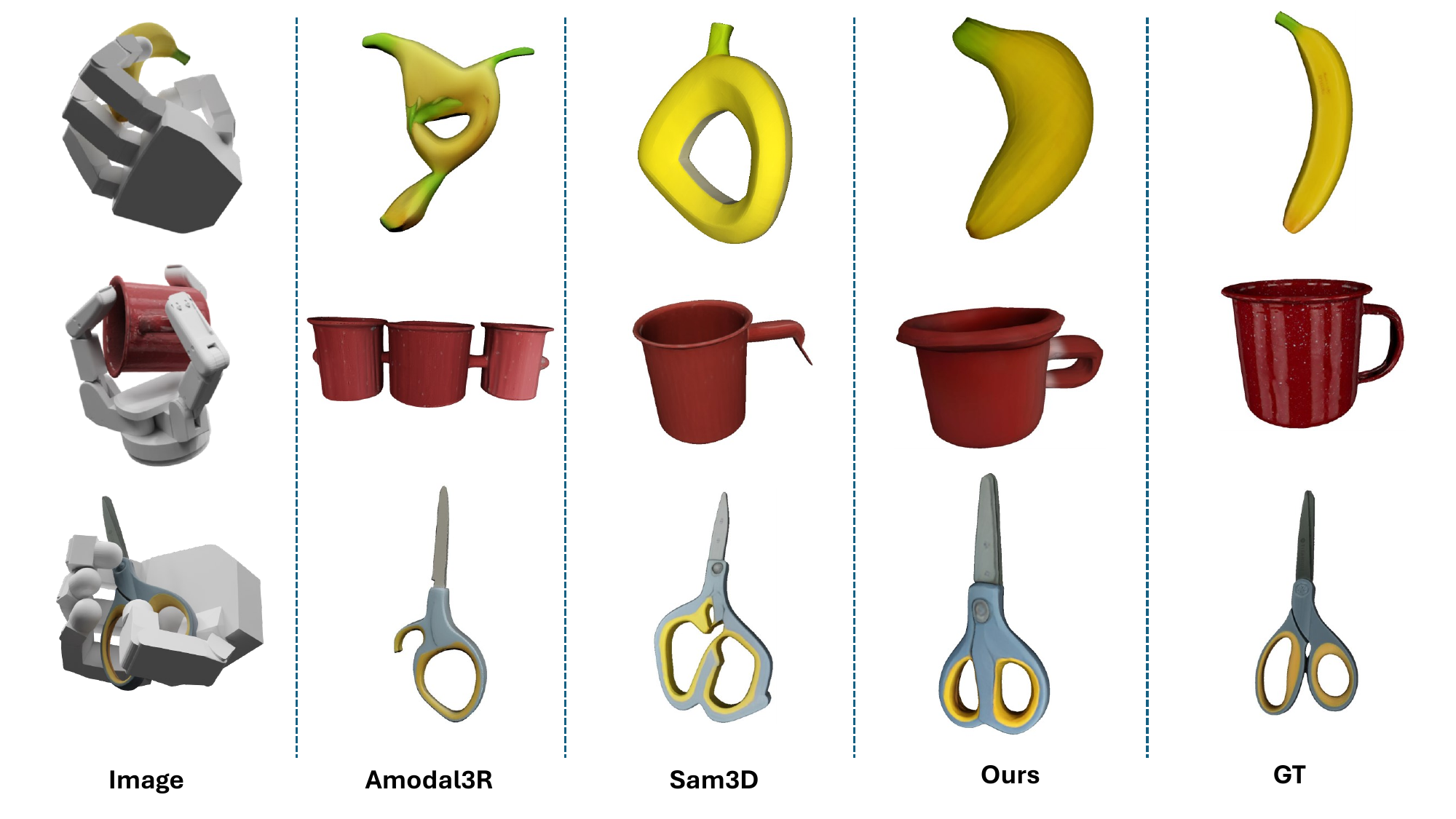}
  \caption{\textbf{Qualitative examples}. Contact cues are critical for recovering feasible shapes under heavy occlusion. Compared to vision-only baselines, integrating touch improves completion of occluded regions and reduces physically implausible artifacts.}
  \label{fig:qual_3}
\end{figure}

\begin{figure}[h]
  \centering
  \includegraphics[width=\textwidth]{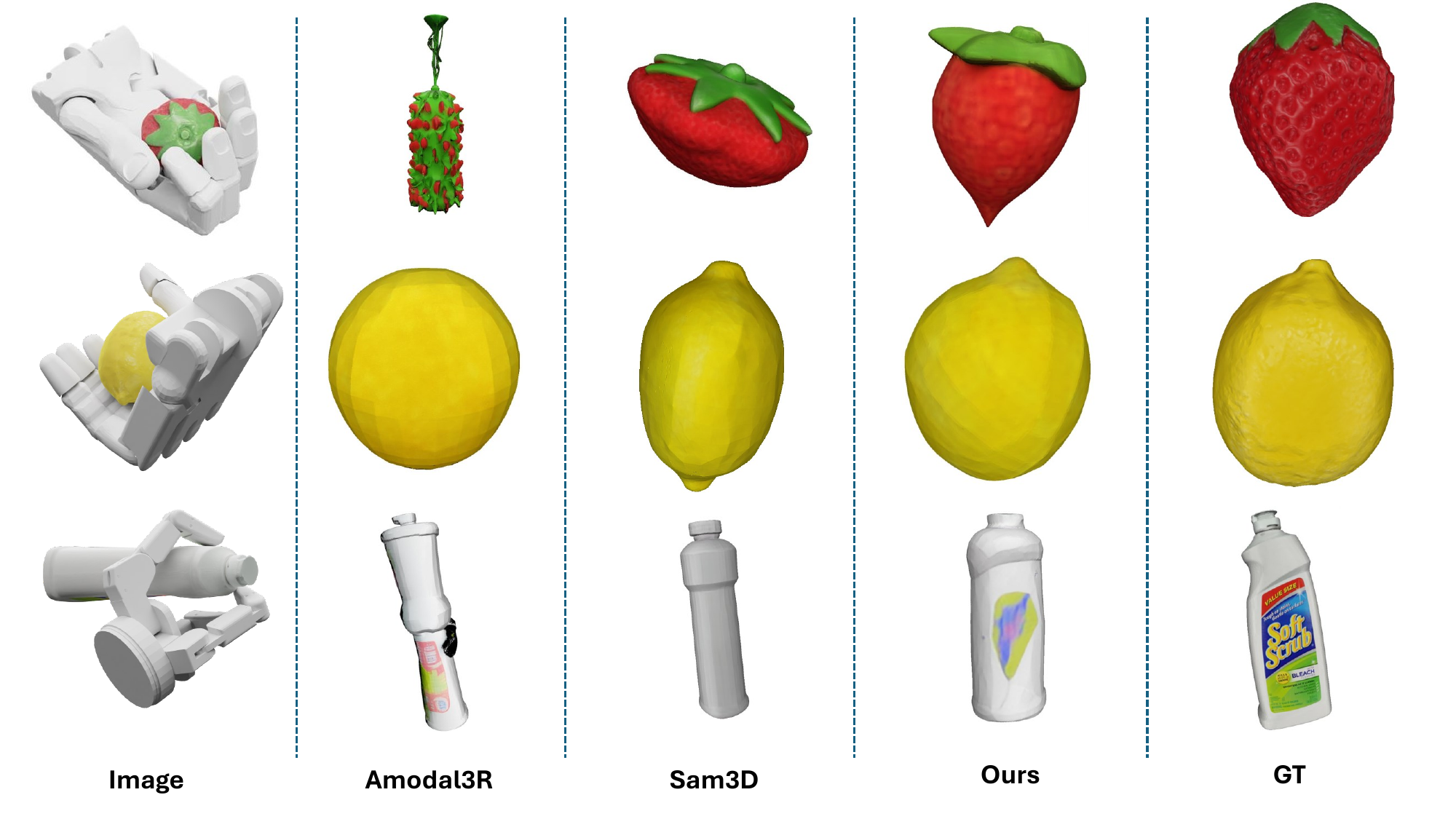}
  \caption{\textbf{Qualitative examples}. Additional qualitative results showing the effect of physical constraints. Across all rows, our method produces higher-fidelity reconstructions and more consistent relative dimensions by enforcing contact consistency and non-interpenetration.}
  \label{fig:qual_4}
\end{figure}

\cref{fig:qual_2,fig:qual_3,fig:qual_4} report additional qualitative results in simulation. In particular, \cref{fig:qual_2} shows reconstructions of the same objects across different end-effectors and camera views, while \cref{fig:qual_3,fig:qual_4} provide further examples in which contact cues have a substantial impact on the recovered geometry.

\end{document}